\documentclass[letterpaper,journal]{IEEEtran}
\usepackage{amsmath,amsfonts}
\usepackage{algorithmic}
\usepackage{algorithm}
\usepackage{array}
\usepackage{subcaption}
\usepackage{textcomp}
\usepackage{stfloats}
\usepackage{url}
\usepackage{verbatim}
\usepackage{graphicx}
\usepackage{multirow}
\usepackage{booktabs}
\usepackage{subcaption}
\usepackage{cite}
\usepackage{hyperref}
\hypersetup{
	colorlinks=true, 
	linkcolor=blue,  
	citecolor=blue, 
	urlcolor=red     
}
\begin{document}

\title{\LARGE \bf
Lightweight Dynamic Modeling of Cable-Driven Continuum Robots Based on Actuation-Space Energy Formulation}

\author{Fangju Yang$^{1,2}$, Hang Yang$^{1}$, Ibrahim Alsarraj$^{1}$, Yuhao Wang$^{1}$, Zhegnqiang Zhang$^{2}$,~\IEEEmembership{Senior Member,~IEEE,} \\Ke Wu$^{1*}$,~\IEEEmembership{Member,~IEEE}
\thanks{Received 10 June 2026; accepted 23 August 2026. This paper was recommended for publication by Editor Yong-Lae Park
upon evaluation of the Associate Editor and Reviewers comments. This work was supported by MBZUAI under Project \#848085. \textit{(Corresponding author: Ke Wu)}}
\thanks{$^{1}$Robotics Department, Mohamed bin Zayed University of Artificial Intelligence(MBZUAI), Abu Dhabi 7909, UAE (e-mail: Ke.Wu@mbzuai.ac.ae).}
\thanks{$^{2}$School of Engineering, Qufu Normal University, Rizhao 276826, China.}
}

\markboth{IEEE ROBOTICS AND AUTOMATION LETTERS,~Vol.~11, No.~6, June~2026}%
{Shell \MakeLowercase{\textit{et al.}}: A Sample Article Using IEEEtran.cls for IEEE Journals}


\maketitle

\begin{abstract}
Cable-driven continuum robots (CDCRs) require accurate, real-time dynamic models for high-speed dynamics prediction or model-based control, making such capability an urgent need. In this paper, we propose the Lightweight Actuation-Space Energy Modeling (LASEM) framework for CDCRs, which formulates actuation potential energy directly in actuation space to enable lightweight yet accurate dynamic modeling. Through a unified variational derivation, the governing dynamics reduce to a single partial differential equation (PDE), requiring only the Euler moment balance while implicitly incorporating the Newton force balance. By also avoiding explicit computation of cable--backbone contact forces, the formulation simplifies the model structure and improves computational efficiency while preserving geometric accuracy and physical consistency. Importantly, the proposed framework for dynamic modeling natively supports both force-input and displacement-input actuation modes, a capability seldom achieved in existing dynamic formulations. Leveraging this lightweight structure, a Galerkin space–time modal discretization with analytical time-domain derivatives of the reduced state further enables an average 62.3\% computational speedup over state-of-the-art real-time dynamic modeling approaches.
\end{abstract}


\section{Introduction}
{C}{able}-driven continuum robots (CDCRs) require accurate, real-time dynamic models to support high-speed motion prediction and model-based control, especially in fast manipulation and dynamic interaction scenarios \cite{della2023model}. Representative scenarios include oscillatory motion and vibration-excitation tests driven by step or sinusoidal tendon inputs \cite{1Renda2014Dynamic}\cite{2tummers2023cosserat}, fast shape reconfiguration under external loading \cite{3li2024real}, vibration-suppressed trajectory tracking \cite{4pierallini2023provably}, and high-speed dynamic trajectory tracking based on accurate dynamic models \cite{caradonna2024model}. In such settings, the strong coupling among inertia, damping, and actuation effects makes empirical or quasi-static models inadequate for accurate dynamic prediction and control \cite{george2018control}. Thus, developing dynamically consistent and computationally efficient models is crucial for advancing CDCRs toward high-performance dynamic operation \cite{7della2023model,8armanini2023soft,9till2019real}.

Early studies based on geometric assumptions introduced the Piecewise Constant Curvature (PCC) empirical model for the kinematic modeling of CDCRs \cite{5PHDchirikjian1992theory,6della2019control}. Subsequent work extended PCC to dynamic modeling by treating segment-level generalized parameters as system states and incorporating equivalent inertia, damping, and actuation terms \cite{2tummers2023cosserat,7della2023model,8armanini2023soft}. However, under significant distributed loads or geometric nonuniformities, PCC-based models lose accuracy in capturing curvature distribution and dynamic response \cite{6della2019control}. Although variable- and polynomial-curvature models improve precision \cite{8armanini2023soft}, they increase computational complexity and the cost of parameter identification \cite{9till2019real}.

Beyond geometric and empirical approaches, data-driven and learning-based methods have also been explored for dynamic modeling and compensation, including iterative learning-based identification \cite{14pierallini2023provably}, Koopman operator embeddings \cite{15bruder2021koopman}, and deep Koopman frameworks for nonlinear dynamics approximation \cite{16komeno2022deep}. While capable of capturing complex loading and coupling effects, these methods are sensitive to data quality and feature representations and often require substantial computational and training resources \cite{chen2024data}. Their generalizability, interpretability, and control suitability still depend on integration with physical priors \cite{15bruder2021koopman,16komeno2022deep}.

In contrast, physics-based dynamic formulations such as FEM \cite{faure2012sofa} and rod theories \cite{11boyer2020dynamics} systematically capture coupled bending, torsion, and axial deformation, providing higher predictive accuracy under complex loading and geometric variations \cite{armanini2023soft}. However, these formulations are strongly nonlinear and high-dimensional, making real-time simulation challenging \cite{armanini2023soft,goury2018fast}. To improve computational efficiency, model reduction and parameterization strategies have been developed, including Model Order Reduction (MOR) \cite{goury2018fast} and strain-based parameterizations \cite{mathew2024reduced}, with recent efforts even enabling real-time differentiable modeling through analytical derivatives in the time domain \cite{mathew2025analytical}.

While physics-based dynamic formulations and reduced-order models have recently achieved real-time 
performance \cite{mathew2025analytical}, their body-frame potential-energy representations, 
Lie-group variational operations, and space–time strain discretizations often result in complex 
formulations that are difficult to implement and generalize, especially for users without a 
specialized background. Motivated by these limitations and inspired by \cite{wu2026lightweight}, we 
streamline the physical modeling for cable-driven actuation and obtain a more compact and tractable 
dynamic formulation. To this end, we propose the Lightweight Actuation-Space Energy Modeling (LASEM) 
framework, which formulates actuation potential energy directly in actuation space and, together 
with the proposed Galerkin-based PDE solver, enables lightweight, accurate, and computationally 
efficient dynamic modeling. The contributions are summarized as follows:

\textbf{Contributions.}
\begin{itemize}
    \item \textbf{Single-PDE dynamic formulation.}
    A unified variational derivation producing a single governing PDE that requires only the Euler moment balance while implicitly satisfying the Newton force balance.

    \item \textbf{No explicit cable--backbone contact-force computation.}
    The actuation-space formulation removes the need for explicit cable--backbone contact-force modeling, simplifying the dynamics and improving computational efficiency.

    \item \textbf{Displacement- and force-input compatibility.}
    Native support for both displacement-input and force-input actuation, a capability rarely achieved in existing dynamic formulations.

    \item \textbf{Broad applicability.}
    A unified modeling interface adaptable to diverse actuation and geometry layouts: arbitrary tendon routings, and nonuniform geometries.

    \item \textbf{Substantial computational speedup.}
Using a Galerkin-based modal discretization yields analytical time-domain derivatives of the reduced state, which in turn enable LASEM to achieve an average 62.3\% speedup over state-of-the-art real-time dynamic models.
\end{itemize}

\section{LASEM Dynamic Modeling Framework}
\begin{figure}
	\vspace*{0.2cm} 
    \centering
    \includegraphics[width=0.9\linewidth]{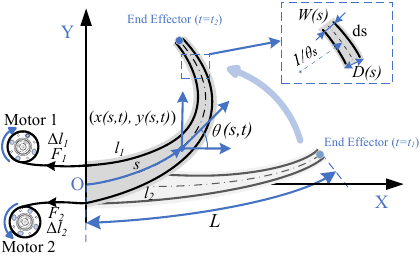}
    \caption{A Cable-driven Continuum Robot}
    \label{fig:placeholder}
\end{figure}
This section details the LASEM dynamic modeling framework for planar continuum robots through a systematic derivation. 
The CDCR illustrated in Fig.~\ref{fig:placeholder} is parameterized as a 
geometrically nonlinear Euler–Bernoulli beam with spatially varying cross-section $D(s)$, second moment of area $I(s)$, 
cross-sectional area $A(s)$, and cable spacing $W(s)$, with two inextensible cables symmetrically 
embedded along its sides. The cable forces $F_1(t)$ and $F_2(t)$ are applied through motors, generating tensile forces along the cables in the pulling directions indicated in Fig.~1, producing
the corresponding cable lengths $l_1(t)$ and $l_2(t)$ along the robot body, with displacements 
$\Delta l_1 = L - l_1$ and $\Delta l_2 = L - l_2$. The arc-length coordinate is $s\in[0,L]$, $L$ denotes the total length of the continuum robot, $\theta(s,t)$ denotes the local bending angle at position $s$, and 
$\theta_s(s,t)$ denotes the curvature. The backbone position in the global frame is 
$(x(s,t),y(s,t))$. Throughout this paper, $(\cdot)_t=\partial(\cdot)/\partial t$, $(\cdot)_{tt}=\partial^2(\cdot)/\partial t^2$, 
$(\cdot)_s=\partial(\cdot)/\partial s$, and $(\cdot)_{ss}=\partial^2(\cdot)/\partial s^2$. 

\subsection{Kinematics and Geometric Constraints}
Under the assumption of inextensibility, the backbone geometry of the studied CDCR is given by
\begin{equation}
\small
\begin{aligned}
    x(s,t)=\int_{0}^{s}\!\cos\theta(\xi,t)\,\mathrm{d}\xi,\ 
y(s,t)=\int_{0}^{s}\!\sin\theta(\xi,t)\,\mathrm{d}\xi.
\end{aligned}
\end{equation}
Following the curvature–tendon displacement relation in \cite{wu2026lightweight}, the  
cable displacement actuation satisfies
\begin{equation}\label{eq_delta_L}
\small
\Delta l(t)=\Delta l_1=-\Delta l_2=\frac{1}{2}\!\int_{0}^{L}W(s)\,\theta_s(s,t)\,\mathrm{d}s,
\end{equation}
which serves as the boundary constraint for the dynamic model derived in the next subsection. Here, $\Delta l(t)$ denotes the cable displacement relative to the initial configuration, not elongation.

\subsection{Derived Dynamic Models through Functional Variation}

\subsubsection{Dynamic Modeling with Force Input}

The system Lagrangian is defined as
\begin{equation}
\small
    \begin{aligned}
	L_F[\theta] &= T_{\mathrm{rot}}+T_x + T_y - U_b - G - F_1\Delta l_1 - F_2\Delta l_2 ,
\end{aligned}\label{111}
\end{equation}
where $T_{\mathrm{rot}}$ is the rotational kinetic energy, $T_x$ and $T_y$ are the translational kinetic energies along the $x$- and $y$-directions, $U_b$ is the bending potential energy, $G$ denotes the gravity term and $F_1\Delta l_1, F_2\Delta l_2$ represent external potential term. Substituting~\eqref{eq_delta_L} into~\eqref{111} yields the compact Lagrangian form
\begin{equation}\label{eq_Lglr}
\small
	L_F[\theta] \!
	= \!T_{\mathrm{rot}}+T_x + T_y  - \!U_b -\! G
	  - \frac{\Delta_F(t)}{2}\!\!\int_0^L\!\!\! \theta_s(s,t)\,W(s)\,\mathrm{d}s {,}
\end{equation}
where $\Delta_F(t)=F_1-F_2$. The energy terms are given by
\begin{equation}\label{eq_all}
\small
\begin{aligned}
	T_{\mathrm{rot}} &= \tfrac{1}{2}\rho \!\int_0^L\! I(s)\,\theta_t^2\,\mathrm{d}s,\ T_x = \tfrac{1}{2}\rho \!\int_0^L\! A(s)
	M_x(s,t)^{2}\mathrm{d}s,\\ 
    T_y &= \tfrac{1}{2}\rho \!\int_0^L\! A(s)
	M_y(s,t)^{2}\mathrm{d}s,\ 
	U_b = \tfrac{1}{2}E \!\int_0^L\! I(s)\,\theta_s^2\,\mathrm{d}s,\\
	G &= \!\!\int_{0}^{L}\!\!\!\! q_x(s)
	\Big[s \!-\!\!\! \int_{0}^{s} \!\!\!\cos \theta(\sigma,t)\, \mathrm{d}\sigma \Big] \mathrm{d}s\!-\!\!\!\int_{0}^{L}\!\!\!\! q_y(s)\!
	\int_{0}^{s} \!\!\!\sin \theta(\sigma,t) \mathrm{d}\sigma\mathrm{d}s,
\end{aligned}
\end{equation}
where
\begin{equation*}
\small
\begin{aligned}
    M_x(s,t) &= \int_0^s\theta_t(\sigma,t)\sin\theta(\sigma,t)\, \mathrm{d}\sigma ,\\
    M_y(s,t) &= \int_0^s\theta_t(\sigma,t)\cos\theta(\sigma,t)\, \mathrm{d}\sigma .
\end{aligned}
\end{equation*}
Finally, the corresponding action functional is
\begin{equation}
\small
	S_F[\theta] = \int_{t_0}^{t_1} L_F[\theta(s,t)]\,\mathrm{d}t .
\end{equation}
According to Hamilton’s principle, the physical motion renders $S_F[\theta]$ stationary, thereby ensuring the balance between kinetic and potential energies. To apply Hamilton’s principle, we evaluate the variation of each energy term in the Lagrangian $L_F[\theta]$ defined in \eqref{111}. The variation of the rotational kinetic energy is
\begin{equation*}
\small
\begin{aligned}
       \int_{t_0}^{t_1} \delta T_{\mathrm{rot}}\,\mathrm{d}t
    = \rho\!\int_{t_0}^{t_1}\!\!\int_0^L\! I(s)\,\theta_t\,\delta\theta_t\,\mathrm{d}s\,\mathrm{d}t, 
\end{aligned}
\end{equation*}
which after integration by parts in time becomes
\begin{equation}\label{eq_Trot}
\small
\begin{aligned}
    \int_{t_0}^{t_1} \delta T_{\mathrm{rot}}\,\mathrm{d}t
    = -\!\int_{t_0}^{t_1}\!\!\!\int_0^L \rho I(s)\,\theta_{tt}\,\delta\theta\,\mathrm{d}s\,\mathrm{d}t.
    \end{aligned}
\end{equation}
For the translational kinetic energy $T_x$, we write
\begin{equation}
\small
\begin{aligned}
   \int_{t_0}^{t_1}\!\delta T_x\,\mathrm{d}t
    = \rho \!\int_{t_0}^{t_1}\!\!\int_0^L\! A(s)M_x(s,t)\,\delta M_x(s,t)\,\mathrm{d}s\,\mathrm{d}t,
\end{aligned}\label{222}
\end{equation}
where
\begin{equation*}
\small
\begin{aligned}
        \delta M_x(s,t)
    = \!\int_0^s\!\!\left[\delta\theta_t(\sigma,t)\sin\theta + \theta_t(\sigma,t)\cos\theta\,\delta\theta \right]\mathrm{d}\sigma.
\end{aligned}
\end{equation*}
By exchanging the order of integration in \eqref{222}, we obtain the compact expression
\begin{equation}\label{eq_T1}
\small
\begin{aligned}
       \int_{t_0}^{t_1} \delta T_x\,\mathrm{d}t
    = \!\int_{t_0}^{t_1}\!\!\int_0^L \rho\sin\theta(\sigma,t)\mathcal{J}_x(\sigma,t)\,\delta\theta(\sigma,t)\mathrm{d}\sigma\mathrm{d}t, 
\end{aligned}
\end{equation}
where
\begin{equation*}
\small
\begin{aligned}
    \mathcal{J}_x(\sigma,t)
    = \!\int_{\sigma}^{L}\!\! A(\xi)
       \int_0^{\xi}\!
       \left[\theta_{tt}(\eta,t)\sin\theta + \theta_t(\eta,t)^2\cos\theta\right]
       \mathrm{d}\eta\,\mathrm{d}\xi.   
\end{aligned}
\end{equation*}
Following a similar derivation to that of $T_x$, the final expression of the variation of $T_y$ is obtained as 
\begin{equation}\label{eq:T2_final}
\small
    \int_{t_0}^{t_1}\!\!\delta T_y\,\mathrm{d}t
    = -\!\int_{t_0}^{t_1}\!\!\!\int_{0}^{L}\!\!\rho\,\cos\theta(\sigma,t)\,
    \mathcal{J}_y(\sigma,t)\,
    \delta\theta(\sigma,t)\,\mathrm{d}\sigma\,\mathrm{d}t,
\end{equation}
where
\begin{equation*}
\small
\begin{aligned}
    \mathcal{J}_y(\sigma,t)\!
    = \!\!\!\int_{\sigma}^{L}\!\!\!\!\!\! A(\xi)\!\!
    \int_{0}^{\xi}\!
    \Big[\theta_{tt}(\eta,t)\cos\theta
    - \theta_t(\eta,t)^2\sin\theta\Big]\!\mathrm{d}\eta\mathrm{d}\xi.
\end{aligned}
\end{equation*}
The variations of the remaining potential terms are obtained as follows.
The variation of the bending potential energy is
\begin{equation*}
\small
\begin{aligned}
        \delta U_b = E \int_0^L I(s)\, \theta_s(s,t)\, \delta\theta_s(s,t)\, \mathrm{d}s.
\end{aligned}
\end{equation*}
Integrating by parts with respect to $s$ gives
\begin{equation}\label{eq_Ub}
\small
    \begin{aligned}
	\delta U_b
	&= \mathcal{B}_1 \!-E \int_0^L \big[I(s)\theta_s(s,t)\big]_s\delta\theta(s,t) \mathrm{d}s\\
	&= \mathcal{B}_1\!-\!\!\!\int_0^L \!\!\!E \Big[ I_s(s)\theta_s(s,t)\! + \!I(s)\theta_{ss}(s,t) \Big]\delta\theta(s,t)\mathrm{d}s,
\end{aligned}
\end{equation}
where
\begin{equation}\label{eq_B1}
\small
\begin{aligned}
    \mathcal{B}_1 = \Big[E\,I(s)\,\theta_s(s,t)\,\delta\theta(s,t)\Big]_{s=0}^{s=L}.
\end{aligned}
\end{equation}
The variation of the external distributed loading term is
\begin{equation}\label{eq_Gx_Gy}
\small
\begin{aligned}
    	\delta G
	=\!\! \int_0^L \!\!\!\!\!\big[Q_x(s)\sin\theta(s,t)\! - Q_y(s)\cos\theta(s,t)\big]\, \!\delta\theta(s,t)\, \mathrm{d}s{,}
\end{aligned}
\end{equation}
where
\begin{equation*}
    \small
    \begin{aligned}
            Q_x(s) &= \int_s^L q_x(\xi)\, \mathrm{d}\xi, \ Q_y(s) = \int_s^L q_y(\xi)\, \mathrm{d}\xi \label{eq:Q}.
    \end{aligned}
\end{equation*}
Next, we evaluate the variation of the actuation term 
\begin{equation}\label{key}
\small
\begin{aligned}
            \delta\big[-\Delta_F(t)\Delta l(t)\big] 
		= -\frac{\Delta_F(t)}{2}\int_0^L \!\!\!W(s)\delta\theta_s(s,t)\mathrm{d}s.
\end{aligned}
\end{equation}
Integrating by parts with respect to $s$ yields
\begin{equation}\label{eq_F}
\small
\begin{aligned}
    \delta\big[-\Delta_F(t)\Delta l(t)\big]
		= \frac{\Delta_F(t)}{2}\int_0^L \!\!\!W_s(s)\delta\theta(s,t)\mathrm{d}s - \mathcal{B}_2,
\end{aligned}
\end{equation}
where
\begin{equation}\label{eq_B3}
\small
\begin{aligned}
    		\mathcal{B}_2 = \Big[\frac{\Delta_F(t)}{2}W(s)\delta\theta(s,t)\Big]_{s=0}^{s=L}.
\end{aligned}
\end{equation}
Collecting the results in 
Eqs.~(\ref{eq_Trot}), (\ref{eq_T1}), (\ref{eq:T2_final}), (\ref{eq_Ub}), (\ref{eq_Gx_Gy}) 
together with~(\ref{eq_F}), and noting that $\delta\theta$ is arbitrary, 
the stationarity condition $\delta S_F = 0$ leads to the governing dynamic equation
\begin{equation}\label{dynamic_F}
\small
	\begin{aligned}
		\rho I(s)\theta_{tt}(s,t)\!
		+ \!\!\rho \mathcal{J}\!(s,t)\!\!=\!\!\big[\!EI(s)\theta_s(s,t)\big]_s\!\!
		+\! \tfrac{\Delta_F(t)}{2} W_s(s)\!
		+\! \mathcal{Q}(s,t).
	\end{aligned}
\end{equation}
The boundary condition follows from the combined boundary terms in 
Eqs.~(\ref{eq_B1}) and~(\ref{eq_B3})
\begin{equation}
    \small
    \begin{aligned}
            \mathcal{B}_1 - \mathcal{B}_2
    = \Big[\big[E I(s)\theta_s(s,t) + \frac{\Delta_F(t)}{2}W(s)\big]\delta\theta(s,t)\Big]_{0}^{L}.
    \end{aligned}
\end{equation}
Since $\delta\theta$ is arbitrary at $s=L$, the natural moment boundary condition is
\begin{equation}
\small
\begin{aligned}
    E I(L)\theta_s(L,t) + \frac{\Delta_F(t)}{2}W(L) = 0.
\end{aligned}
\end{equation}
Integrating Eq.~(\ref{dynamic_F}) with respect to $s$ from $0$ to $L$ incorporates the above 
boundary condition into the dynamic formulation
\begin{equation}\label{dynamic_in_F}
\small
	\begin{aligned}
		\rho \!\!\int_{0}^{L} \!\!\!\!\!I(s)\theta_{tt}(s,t)\mathrm{d}s
		\!+ \!\rho \!\!\int_{0}^{L} \!\!\!\!\!\mathcal{J}(s,t)\mathrm{d}s
		= \!-\frac{\Delta_F(t)}{2}W(0)
		\!+ \!\!\!\int_{0}^{L} \!\!\!\!\!\mathcal{Q}(s,t)\mathrm{d}s,
	\end{aligned}
\end{equation}
where
\begin{equation}\label{eq_J_Q}
\small
	\begin{aligned}
		\mathcal{J}(s,t) &= \sin\theta(s,t)\mathcal{J}_x(s,t) + \cos\theta(s,t)\mathcal{J}_y(s,t),\\
        \mathcal{J}_x(s,t)\!
        &= \!\!\!\int_{s}^{L} \!\!\!\!\!\!A(\xi)\!\!\!
        \int_{0}^{\xi}\!\!
        \Big[\theta_{tt}(\eta,t)\sin\theta(\eta,t)\!+\!\theta_t(\eta,t)^2\cos\theta(\eta,t)\Big]\!\mathrm{d}\eta\mathrm{d}\xi,\\
        \mathcal{J}_y(s,t)\!
        &= \!\!\!\int_{s}^{L}\!\!\!\!\!\! A(\xi)\!\!\!
        \int_{0}^{\xi}\!\!
        \Big[\theta_{tt}(\eta,t)\cos\theta(\eta,t)\!
        - \!\theta_t(\eta,t)^2\sin\theta(\eta,t)\Big]\!\mathrm{d}\eta \mathrm{d}\xi,\\
		\mathcal{Q}(s,t) &= Q_y(s)\cos\theta(s,t) - Q_x(s)\sin\theta(s,t).
	\end{aligned}
\end{equation}
The initial and boundary conditions are
\begin{equation}
\small
    \begin{aligned}
	\theta(s, 0) = 0,\ \theta_t(s, 0) = 0,\ \theta(0, t) = 0.
    \end{aligned}
\end{equation}

\subsubsection{Dynamic Modeling with Displacement Inputs}
In most existing dynamic formulations for CDCRs, cable tension forces are used as the
actuation inputs \cite{della2023model}. In practice, however, true force control is difficult because
accurate tendon force sensing requires inline load cells \cite{bajo2016hybrid}, and estimating tension
from motor current is unreliable due to gearbox friction \cite{gonzalez2025evaluation}. As a result,
many controllers directly command cable displacements while assuming a monotonic displacement–tension
relationship \cite{wu2022fem1,li2022equivalent}, which does not constitute actual force control and
limits the benefits of dynamic modeling. To address this gap, we introduce a displacement-input
formulation that treats the cable displacement variation $\Delta l(t)$ as the actuation variable.
The corresponding Lagrangian is

\begin{equation}\label{eq_Lg_L}
\small
    \begin{aligned}
	L_D[\theta] &= T_{\mathrm{rot}}+T_x + T_y- U_b - G, 
\end{aligned}
\end{equation}
\hspace{-0.1em}where $T_{\mathrm{rot}}$, $T_x$, $T_y$, $U_b$, and $G$ are defined in the same manner as in Eq.~(\ref{eq_all}), 
and $\Delta l(t)$ still satisfies the constraint~(\ref{eq_delta_L}). The action functional is
\begin{equation}
\small
\begin{aligned}
    	S_D[\theta(s,t)] = \int_{t_0}^{t_1} L_D[\theta(s,t)] \, \mathrm{d}t.
\end{aligned}
\end{equation}
Incorporating the constraint~(\ref{eq_delta_L}), the variational problem becomes
\begin{equation}\label{eq_constraint}
\small
 	\begin{aligned}
			 	\theta(s,t) = &\arg\min\int_{t_0}^{t_1}L_D[\theta(s,t)] \mathrm{d}t, \\ 
			&\text{s.t.} (\ref{eq_delta_L}). 
		\end{aligned}
 \end{equation}
\hspace{-0.25em}To impose the constraint, we introduce a time-dependent Lagrange multiplier $\lambda(t)$ and define the augmented functional
\begin{equation}
\small
\begin{aligned}
    \mathcal{H}_D(\theta,\lambda) 
	= \int_{t_0}^{t_1} \Big[ 
	L_D[\theta(s,t)] 
	- \Psi(t) 
	\Big] \mathrm{d}t,
\end{aligned}
\end{equation}
where
\begin{equation*}
\small
\begin{aligned}
        \Psi(t) = \lambda(t) 
	\Big[ \Delta l(t) - \frac{1}{2} \int_0^L \theta_s(s,t) W(s)\, \mathrm{d}s \Big]. 
\end{aligned}
\end{equation*}
\hspace{-0.1em}Variation with respect to $\lambda(t)$ enforces the constraint, while variation with respect to
$\theta(s,t)$ yields the dynamic equation. Taking the variation with respect to $\lambda(t)$ gives
\begin{equation}
    \small
    \begin{aligned}
        \delta_\lambda \mathcal{H}_D 
        = -\!\!\!\int_{t_0}^{t_1} 
        \!\!\Big[
        \Delta l(t) - \frac{1}{2} \int_{0}^{L}\!\!\! \theta_s(s,t) W(s)\, \mathrm{d}s 
        \Big] 
        \delta\lambda(t)\, \mathrm{d}t.
    \end{aligned}
\end{equation}
Implying~(\ref{eq_delta_L}) since $\delta\lambda(t)$ is arbitrary. The variation of the multiplier term is
\begin{equation}\label{eq_Psi}
\small
    \begin{aligned}
	\delta \Psi 
	&= -\frac{1}{2}\!\int_{t_0}^{t_1} \!\!\!\lambda(t)
	\Big[\int_{0}^{L} W(s)\delta\theta_s(s,t)\mathrm{d}s \Big] \mathrm{d}t \\
	&= -\int_{t_0}^{t_1} \!\!\!\mathcal{B}_3 \mathrm{d}t
	+ \frac{1}{2}\int_{t_0}^{t_1} \!\!\!\lambda(t)
	\Big[\int_{0}^{L} W_s(s)\delta\theta(s,t)\mathrm{d}s \Big] \mathrm{d}t,
\end{aligned}
\end{equation}
where the boundary term is
\begin{equation}\label{eq_B2}
\small
\begin{aligned}
    	\mathcal{B}_3 = \Big[\tfrac{\lambda(t)}{2}W(s)\,\delta\theta(s,t)\Big]_{0}^{L}.
\end{aligned}
\end{equation}
Collecting the variations in 
Eqs.~(\ref{eq_Trot}), (\ref{eq_T1}), (\ref{eq:T2_final}), (\ref{eq_Ub}),
(\ref{eq_Gx_Gy}), and~(\ref{eq_Psi}), and enforcing $\delta S_D = 0$, we obtain
\begin{equation}\label{dynamic_1}
\small
	\begin{aligned}
		\rho I(s)\theta_{tt}(s,t)\!
		+ \!\rho \mathcal{J}(s,t)\!=\!\Big[EI(s)\theta_s(s,t)\Big]_s\!\!\! \! 
		-\! \frac{\lambda(t)}{2} W_s(s)\!
		+\! \mathcal{Q}(s,t),
	\end{aligned}
\end{equation}
The boundary contributions from Eqs.~(\ref{eq_B1}) and~(\ref{eq_B2}) give
\begin{equation*}
    \small
    \begin{aligned}
            \mathcal{B}_1 - \mathcal{B}_3
    = \Big[\big[E I(s)\theta_s(s,t)-\tfrac{\lambda(t)}{2}W(s)\big]\delta\theta(s,t)\Big]_{0}^{L}.
    \end{aligned}
\end{equation*}
Since $\delta\theta$ is arbitrary at $s=L$, the boundary condition is
\begin{equation}\label{eq_lamber}
\small
\begin{aligned}
    	E I(L)\theta_s(L,t) - \tfrac{\lambda(t)}{2}W(L) = 0.
\end{aligned}
\end{equation}
Solving~(\ref{eq_lamber}) yields
\begin{equation}\label{lamber_value}
\small
\begin{aligned}
    	\lambda(t) = \frac{2E I(L)\theta_s(L,t)}{W(L)}.
\end{aligned}
\end{equation}
Using the constraint~(\ref{eq_delta_L}) and integrating by parts,
\begin{equation}\label{eq_deltaL}
\small
	\begin{aligned}
		\Delta l(t) = \frac{1}{2}W(L)\theta(L,t) - \frac{1}{2}\int_{0}^{L}W_s(s)\theta(s,t)\mathrm{d}s.
	\end{aligned}
\end{equation}
We can get
\begin{equation}\label{lamber_value_fin}
\small
\begin{aligned}
    	\lambda(t) = \frac{2E I(L)\theta_s(L,t)\theta(L,t)}{2\Delta l(t) + \int_{0}^{L}W_s(s)\theta(s,t)\mathrm{d}s}.
\end{aligned}
\end{equation}
Finally, integrating Eq.~(\ref{dynamic_1}) from $s=0$ to $s=L$ yields
\begin{equation}\label{dynamic_in}
\small
	\begin{aligned}
		\rho \!\!\int_{0}^{L}\!\!\!I(s)\theta_{tt}(s,t)\mathrm{d}s
		+ \rho\!\!\int_{0}^{L}\mathcal{J}(s,t)\mathrm{d}s
		= \frac{\lambda(t)}{2}W(0)
		+\!\! \int_{0}^{L}\!\!\!\mathcal{Q}(s,t)\mathrm{d}s.
	\end{aligned}
\end{equation}
\subsubsection{Unified Dynamic Formulation}
{It is worth noting that Eqs. \eqref{dynamic_in_F} and \eqref{dynamic_in} share an identical dynamic structure. So, we further establish a unified dynamic framework
\begin{equation}
    \small
    \begin{aligned}
        \rho \!\!\int_{0}^{L}\!\!\!I(s)\theta_{tt}(s,t)\mathrm{d}s
		+ \rho\!\!\int_{0}^{L}\mathcal{J}(s,t)\mathrm{d}s
		= \Gamma(t) W(0)
		+\!\! \int_{0}^{L}\!\!\!\mathcal{Q}(s,t)\mathrm{d}s,
    \end{aligned}
\end{equation}
where $\mathcal{J}(s,t)$, $\mathcal{J}_x(s,t)$, $\mathcal{J}_y(s,t)$, and $\mathcal{Q}(s,t)$ follow the previous definitions \eqref{eq_J_Q} and $\Gamma(t)$ is defined as follows
\begin{equation*}
    \small
    \begin{aligned}
        \Gamma(t_n)=
    \begin{cases}
        -\frac{1}{2}\Delta_F(t_n), & \text{Force-Input},\\
        +\frac{1}{2}\lambda(t_n), & \text{Displacement-Input}.
    \end{cases}
    \end{aligned}
\end{equation*}
}
\begin{algorithm}[h]
	\caption{Analytical Derivatives via Galerkin Method}\footnotesize
	\label{alg:galerkin_solver}
	\begin{algorithmic}[1]
        \STATE \textbf{Inputs:} Actuation $\Gamma(t)$ (force or displacement), distributed load $\mathcal Q(s)$, initial states $\mathbf{c}_0$, $(\mathbf{c}_0)_t$.
		\STATE \textbf{Initialization:} Precompute $\mathbf{M}$ and $\!\int_0^L\!\mathbf{\Phi}(s)\mathrm{d}s$.
		\FOR{$k = 1$ to $N$}
		\STATE \textbf{Update:} From $[\mathbf{c}_{k-1}, ({\mathbf{c}}_{k-1})_t]$ evaluate $\mathbf{K}$, $\mathbf{h}$, $\mathbf{f}_\mathbf{Q}$, and $\mathbf{f}_\mathbf{l}(t_k)$.
		\STATE \textbf{Solve:} 
		$\displaystyle({\mathbf{c}}_k)_{tt}=\big[\mathbf{M}+\rho\mathbf{K}(\mathbf{c}_{k-1})\big]^{-1}
		\!\Big(\mathbf{f}_\mathbf{Q}-\rho\mathbf{h}+\mathbf{f}_\mathbf{l}\Big)$.
        \STATE \textbf{Update:} 
        Update the generalized coordinates $\mathbf{c}_k$ and velocities $({\mathbf{c}}_k)_t$; 
        reconstruct the configuration $\theta(s,t_k)$ and end-effector position $(x,y)$.
		\ENDFOR
	\end{algorithmic}
\end{algorithm}

\section{Derivation of Analytical Derivatives via Galerkin Method}
To enable fast numerical computation, the partial differential equation under LASEM framework is 
projected onto a finite-dimensional coefficient space using the Galerkin method 
\cite{sadati2017control}. Let $\{\phi_j(s)\}_{j=0}^{m-1}$ denote the chosen basis functions, arranged as
\begin{equation}
\small
\begin{aligned}
    \mathbf{\Phi}(s)\!\!=\!\![\phi_0(s),\ldots,\phi_{m-1}(s)]^{\mathbf{T}}\!\!,\quad\!\!\!\!\!\!\!\mathbf c_n\!\!=\!\![c_0(t_n),\ldots,c_{m-1}(t_n)]^\mathbf{T}\!\!\!, 
\end{aligned}
\end{equation}
\hspace{-0.5em}where \(t_n = n\Delta t\) are discrete time nodes, and $c_n$ denotes the vector of modal coefficients at time step $t_n$. Under this modal representation, the bending angle and its derivatives at step $n$ are
\begin{equation}\label{eq:def_theta_discrete}
\small
\begin{aligned}
    \theta(s,n)\!=\!\mathbf{\Phi}^{\!\mathbf{T}}\!(s)\mathbf c_n,
    \theta_t(s,n)\!=\mathbf{\Phi}^{\!\mathbf{T}}\!(s)({\mathbf c}_n)_t,
    \theta_{tt}(s,n)\!=\mathbf{\Phi}^{\!\mathbf{T}}\!(s)({\mathbf c}_n)_{tt}.
\end{aligned}
\end{equation}
Substituting~\eqref{eq:def_theta_discrete} into the dynamic equation and enforcing the Galerkin 
orthogonality condition with respect to each basis $\phi_i(s)$ yields the discretized dynamics. At 
time step $n$, the system equation becomes
\begin{equation}\label{eq:disc_main}
\small
\begin{aligned}
    \big[\mathbf{M}\! + \!\rho\mathbf{K}(\mathbf{c}_{n-1})\!\big]\!({\mathbf{c}}_n)_{tt}
	\!= \!\mathbf{f}_\mathbf{Q}(\mathbf{c}_{n-1})\!
	- \!\!\rho\,\mathbf{h}[\mathbf{c}_{n-1},({\mathbf{c}}_{n-1})_{t}]\!
	+ \!\mathbf{f}_\mathbf{l}(t_{n}),
\end{aligned}
\end{equation}
where the system matrices and vectors are
\begin{equation*}\label{eq:fl_vector}
\small
	\begin{aligned}
        \mathbf{M} &= \rho\int_{0}^{L}I(s)\mathbf{\Phi}(s)\mathbf{\Phi}^{\mathbf{T}}(s)\mathrm{d}s,\\
        \mathbf{K}(\mathbf{c}_{n}) &= \!\!\int_{0}^{L}\!\!\Big[\sin\theta(s,n)\Xi^{(\sin)}(s,n)+\cos\theta(s,n)\Xi^{(\cos)}(s,n)\Big]\mathrm{d}s,\\
        \mathbf{f}_\mathbf{Q}(\mathbf{c}_{n}) &=\!\! \int_{0}^{L}\!\!\mathbf{\Phi}(s)\big[Q_{y}(s)\cos\theta(s,n)-Q_{x}(s)\sin\theta(s,n)\big]\mathrm{d}s,\\
        \mathbf{f}_\mathbf{l}(t_{n}) &= \Gamma(t_{n})W(0)\int_{0}^{L}\mathbf{\Phi}(s)\mathrm{d}s. 
    \end{aligned}
\end{equation*}
The intermediate coupling terms are
\begin{equation*}
\small
\begin{aligned}
    \Xi^{(\sin)}(s,n)&=\Big[\int_{s}^{L}A(\xi)\int_{0}^{\xi}\mathbf{\Phi}(\varpi)\sin\theta(\varpi,n)\mathrm{d}\varpi~\mathrm{d}\xi\Big]\Phi^{\mathbf{T}}(s),\\
    \Xi^{(\cos)}(s,n)&=\Big[\int_{s}^{L}A(\xi)\int_{0}^{\xi}\mathbf{\Phi}(\varpi)\cos\theta(\varpi,n)\mathrm{d}\varpi~\mathrm{d}\xi\Big]\Phi^{\mathbf{T}}(s).
\end{aligned}
\end{equation*}
The Coriolis and centrifugal vector is
\begin{equation*}
    \small
    \begin{aligned}
    \mathbf{h}[\mathbf{c}_n,({\mathbf{c}}_n)_{t}]\!\!&=\!\!\!\int_{0}^{L}\!\!\!\!\!\!\mathbf{\Phi}(s)\!\Bigg[\!\sin\theta(s,n)\!\!\!\int_{s}^{L}\!\!\!\!\!\!A(\xi)\!\!\!\int_{0}^{\xi}\!\!\!\!\theta_{t}^{2}(\varpi,n)\cos\theta(\varpi,n)\mathrm{d}\varpi\mathrm{d}\xi  \\
    &\quad -\cos\theta(s,n)\!\!\!\int_{s}^{L}\!\!\!\!\!\!A(\xi)\!\!\!\int_{0}^{\xi}\!\!\!\theta_{t}^{2}(\varpi,n)\sin\theta(\varpi,n)\mathrm{d}\varpi\mathrm{d}\xi\Bigg]\mathrm{d}s. 
\end{aligned}
\end{equation*}
Based on this unified actuation treatment, the accelerated computational framework is implemented 
via an explicit temporal discretization strategy, summarized in Algorithm~\ref{alg:galerkin_solver}.

\begin{table}[!t]
\setlength{\tabcolsep}{6pt} 
\vspace*{0.3cm}
\caption{Simulation and Experimental Parameters}\footnotesize
\centering
\begin{tabular}{lccc}
\hline
\textbf{Parameter} & \textbf{Symbol} & \textbf{Value} & \textbf{Unit} \\
\hline
Manipulator length          & $L$ & 0.40 & m \\
Cross-sectional diameter    & $D$ & 0.004 & m \\
Young’s modulus             & $E$ & $2\times10^{9}$ & Pa \\
Cable spacing               & $W$ & 0.11 & m \\
Second moment of area       & $I$ & $1.26\times10^{-11}$ & m$^{4}$ \\
Cross-sectional area        & $A$ & $1.26\times10^{-5}$ & m$^{2}$ \\
Distributed gravity load    & $q_y;\,q_x$ & $1.4794;\,0$ & N/m \\
Viscous damping coefficient & $c$ & 16 & N·m·s/rad \\
\hline
\end{tabular}
\label{tab:exp_sim_params}
\caption*{\scriptsize{\textbf{Note:}
Viscous damping in CDCR dynamics is commonly modeled as $c\,\dot{\theta}$ \cite{goury2018fast}, representing the dominant effect of frictional energy dissipation.} 
LASEM adopts this standard form by adding $c\,\dot{\theta}$ to \eqref{dynamic_in_F} and \eqref{dynamic_in}, 
with the coefficient $c$ identified from the dissipative behavior of the Cosserat-rod formulations to ensure 
consistent damping across formulations.}
\end{table}
\section{Numerical Simulation and Model Validation}
To evaluate the accuracy and computational efficiency of the proposed LASEM framework, numerical
simulations were conducted under diverse conditions. Within LASEM, two solution strategies were
examined: a conventional spatial discretization (SD) scheme \cite{leveque2007finite}, which is used as a baseline for comparison, and the proposed Galerkin formulation \eqref{eq:disc_main}, which ends up analytical time-domain derivatives. 
This internal comparison highlights the advantages of the Galerkin-based solver for the PDE derived
in LASEM. The framework is further benchmarked against two established Cosserat-rod formulations,
the Geometric Variable Strain (GVS) method \cite{mathew2024reduced} and its differentiable extension
D-GVS \cite{mathew2025analytical}, under identical inputs and boundary conditions.

\begin{figure*}[!t]
	\vspace*{0.15cm} 
    \centering
    \begin{subfigure}{0.245\textwidth}
        \includegraphics[width=\linewidth]{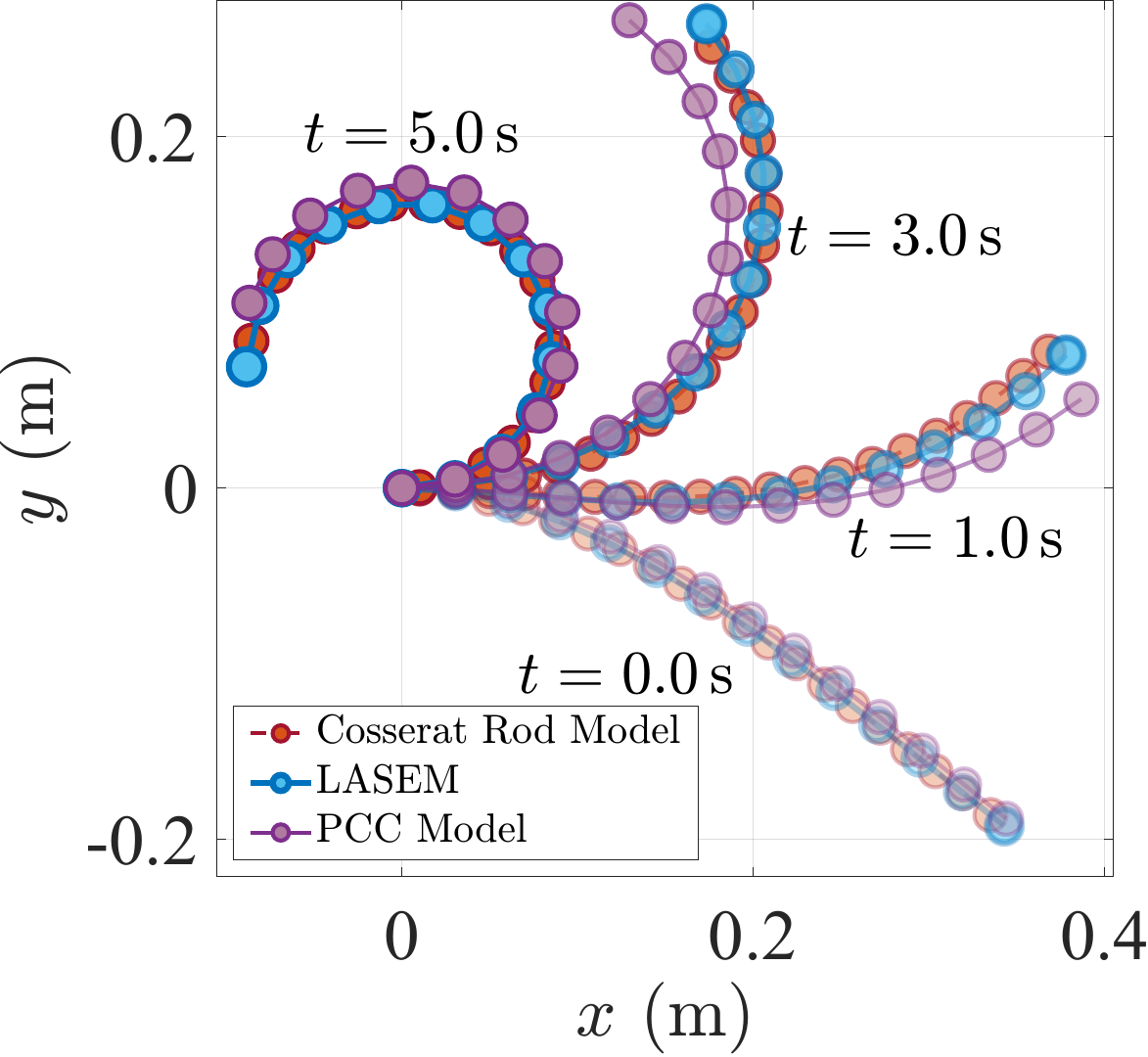}
        \caption{{\footnotesize{$\Delta_F(t)=t$}}}\label{F_con_linear}
    \end{subfigure}
    \begin{subfigure}{0.245\textwidth}
        \includegraphics[width=\linewidth]{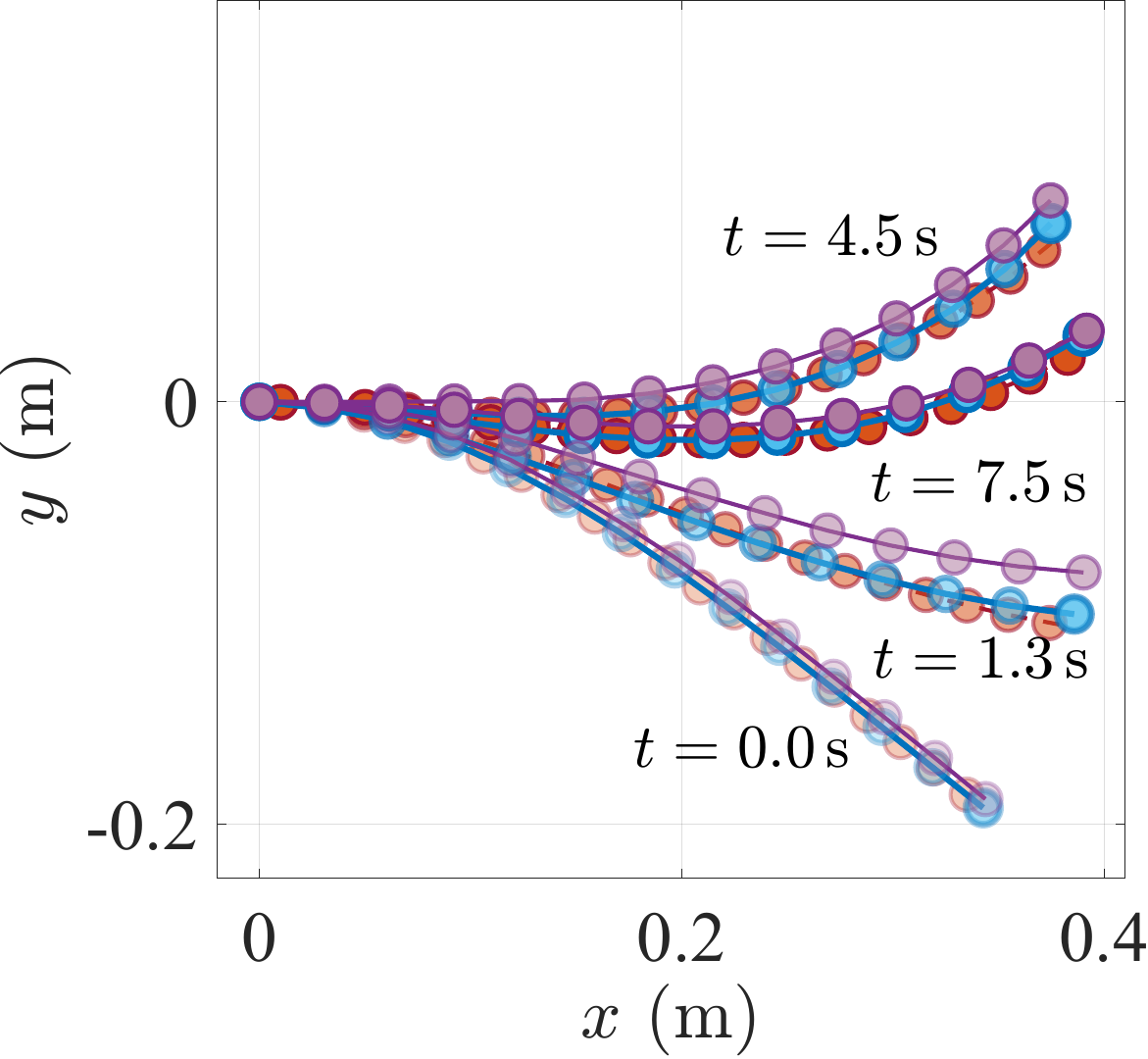}
        \caption{{\footnotesize{$\Delta_F(t)\!\!=\!1.5\!-\!0.3\sin[2\pi(t-1)]$}}}
    \end{subfigure}\label{F_con_sin}
    \begin{subfigure}{0.245\textwidth}
        \includegraphics[width=\linewidth]{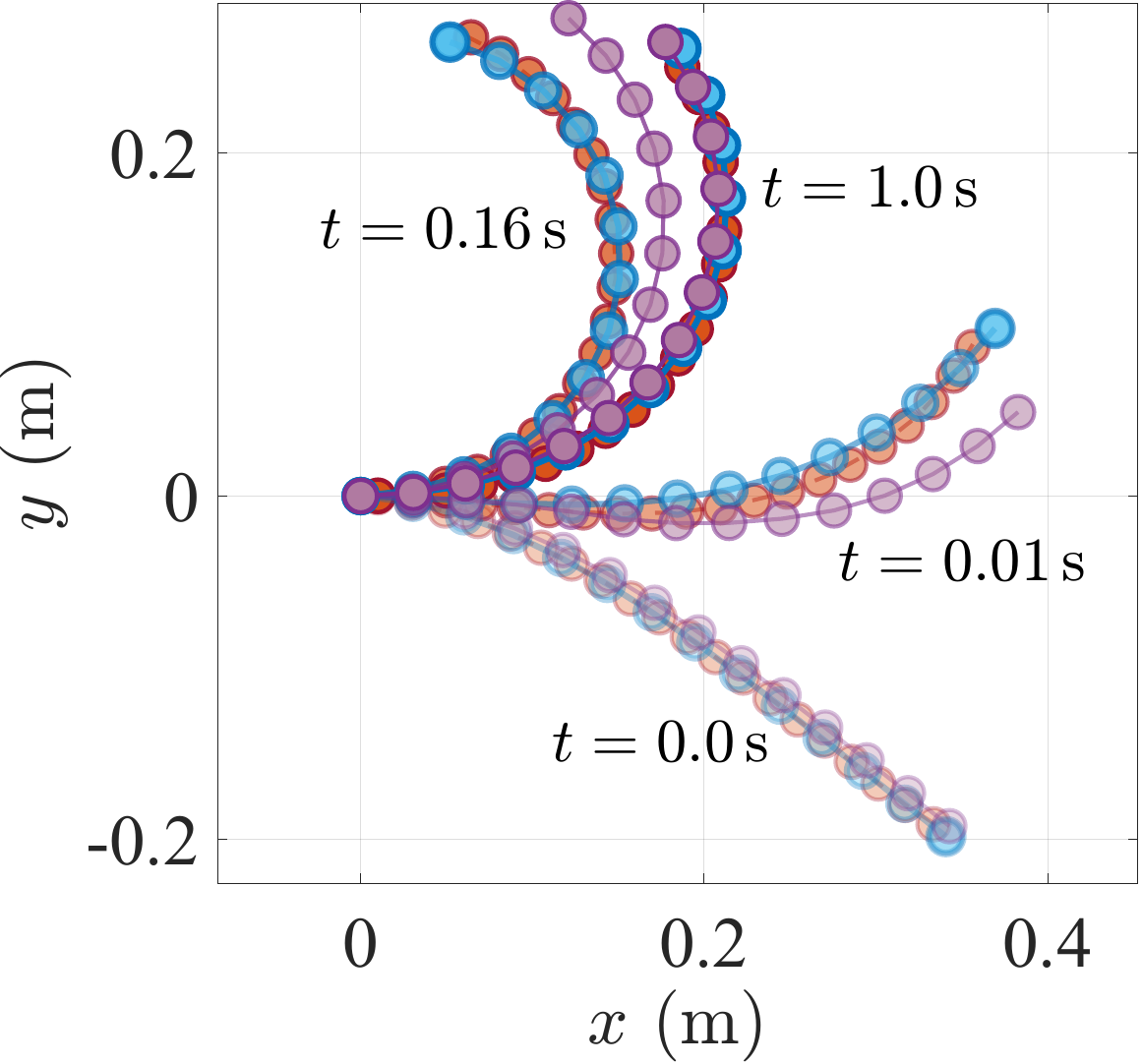}
        \caption{{\footnotesize{$\Delta_F(t)=3u(t)$}}}\label{F_con_step}
    \end{subfigure}
    \begin{subfigure}{0.245\textwidth}
        \includegraphics[width=\linewidth]{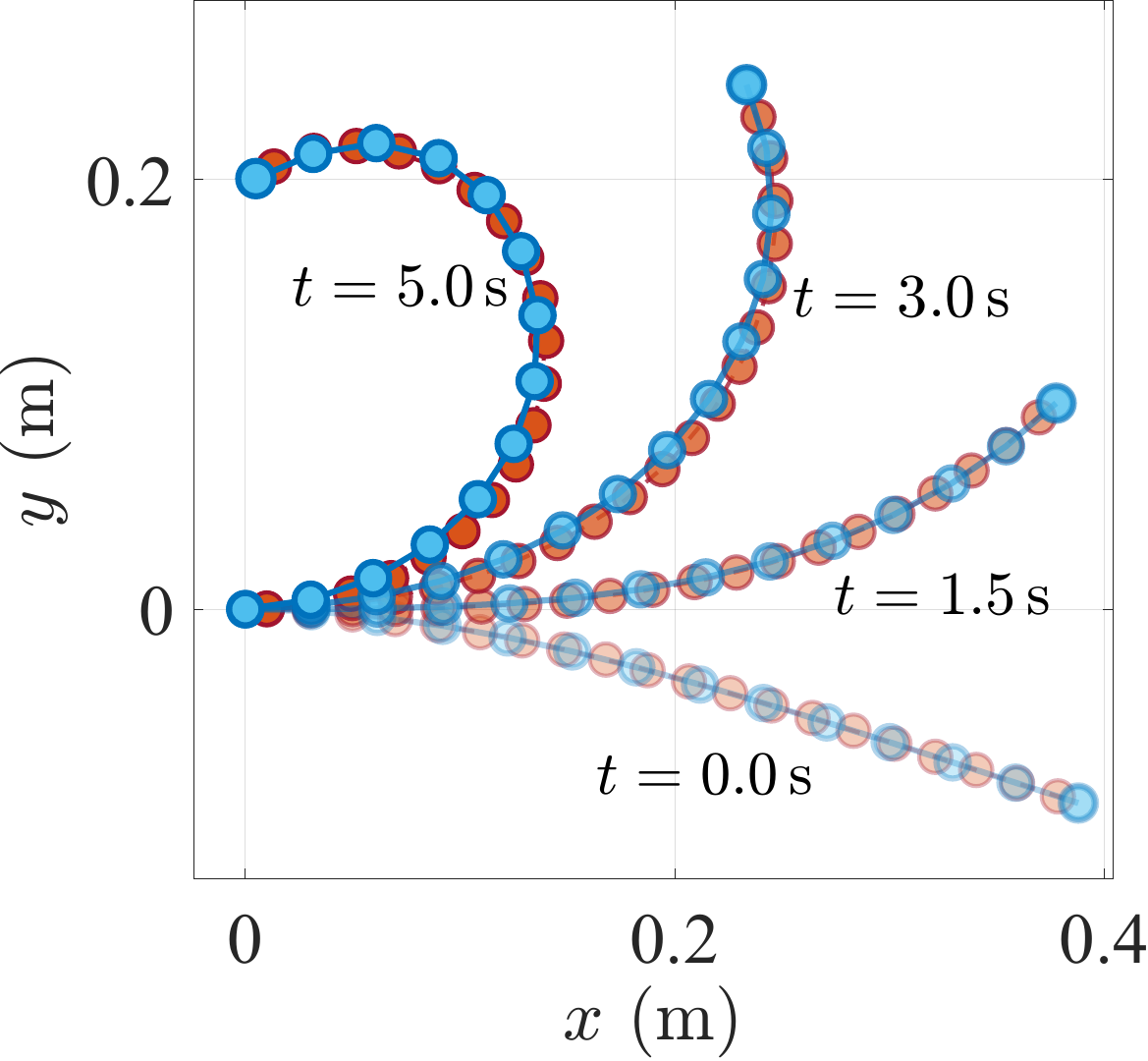}
        \caption{\footnotesize{$\Delta_F(t)=2.75t$}}\label{F_Ds_linear}
    \end{subfigure}
    \begin{subfigure}{0.245\textwidth}
        \includegraphics[width=\linewidth]{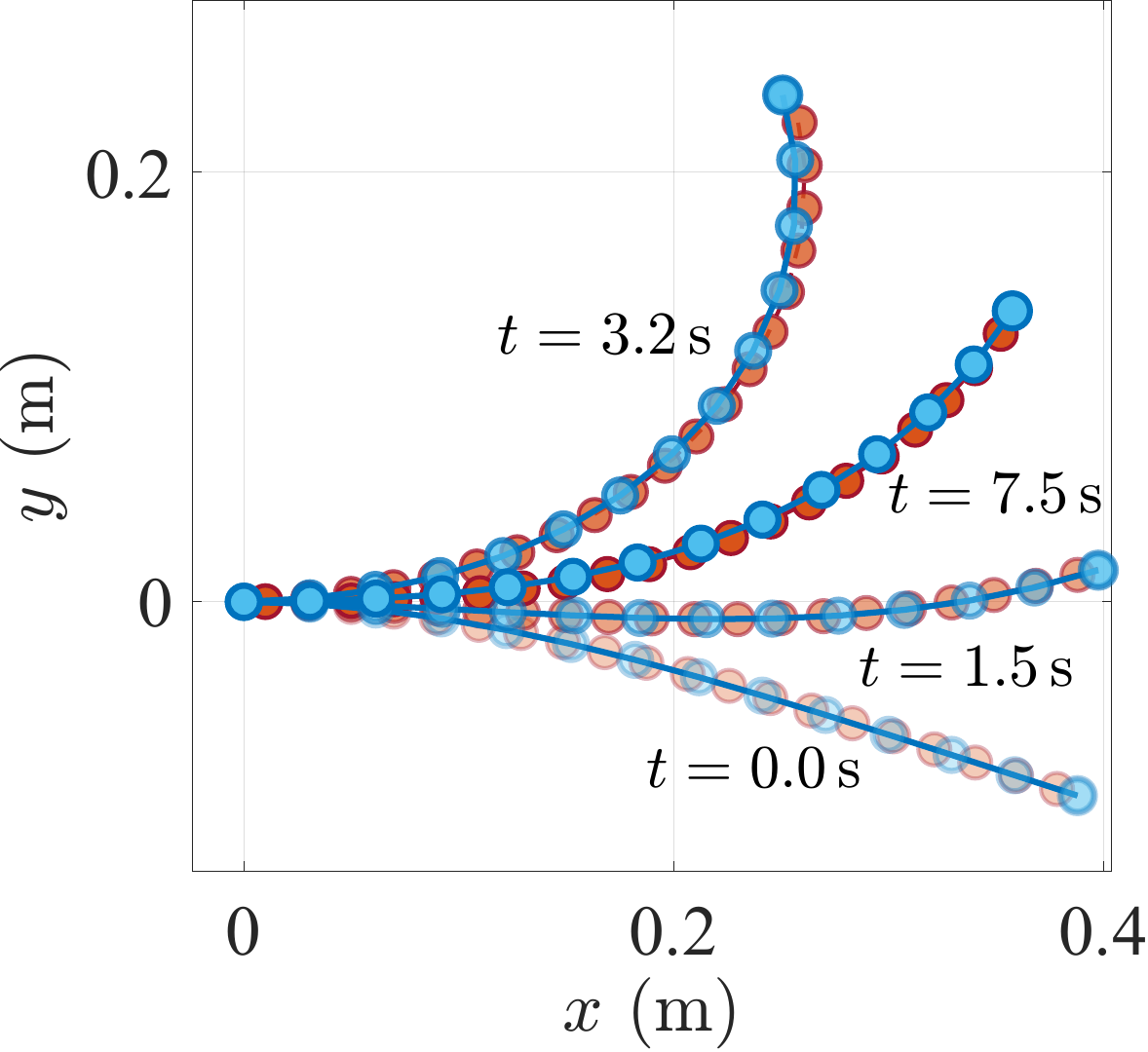}
        \caption{\footnotesize{$\Delta_F(t)\!\!=\!5\!-\!3\sin[2\pi(t-1)]$}}
    \end{subfigure}\label{F_Ds_sin}
    \begin{subfigure}{0.245\textwidth}
 \includegraphics[width=\linewidth]{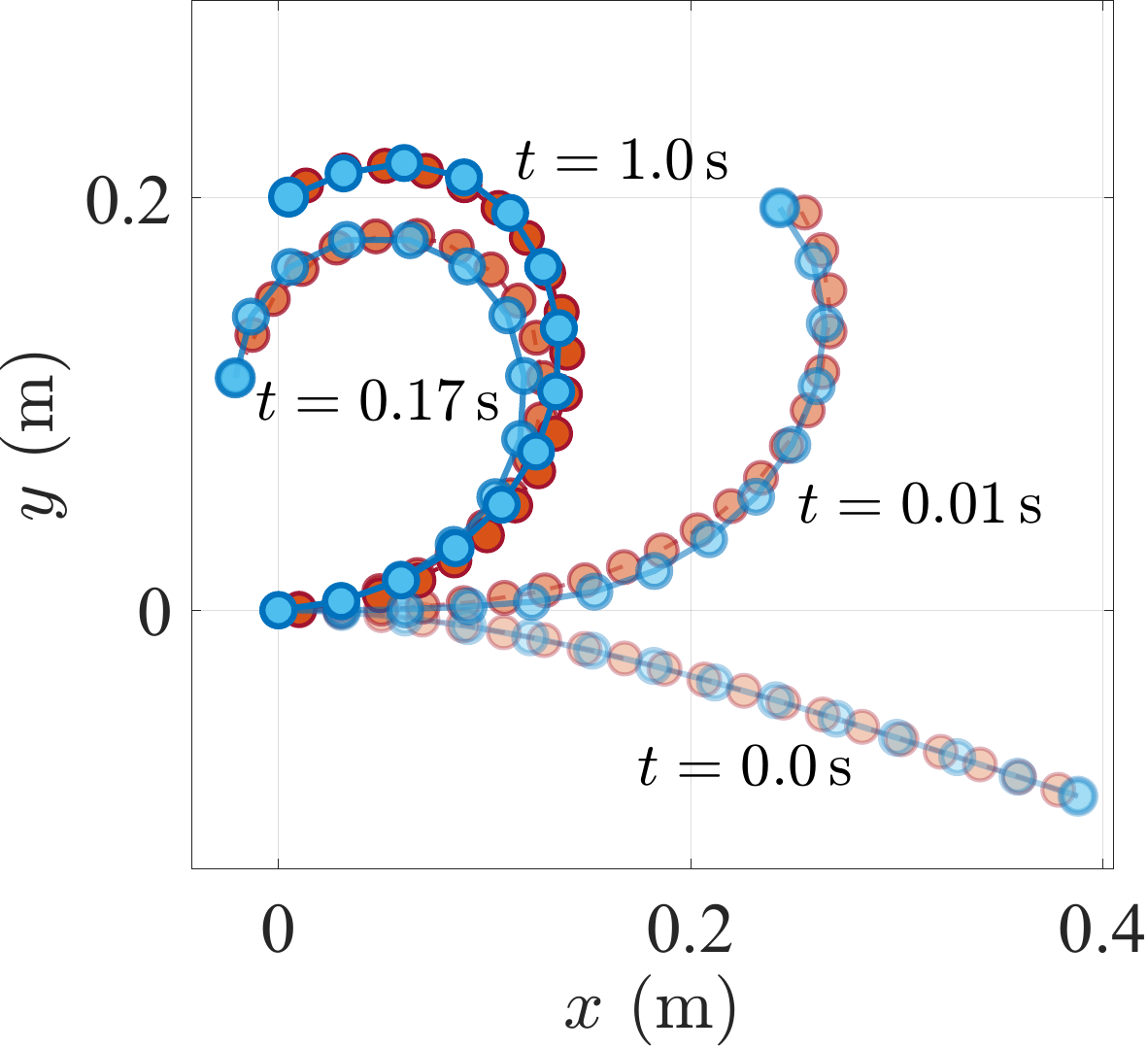}
        \caption{\footnotesize{$\Delta_F(t)=13.75u(t)$}}\label{F_Ds_step}
    \end{subfigure}
    \begin{subfigure}{0.245\textwidth}
\includegraphics[width=\linewidth]{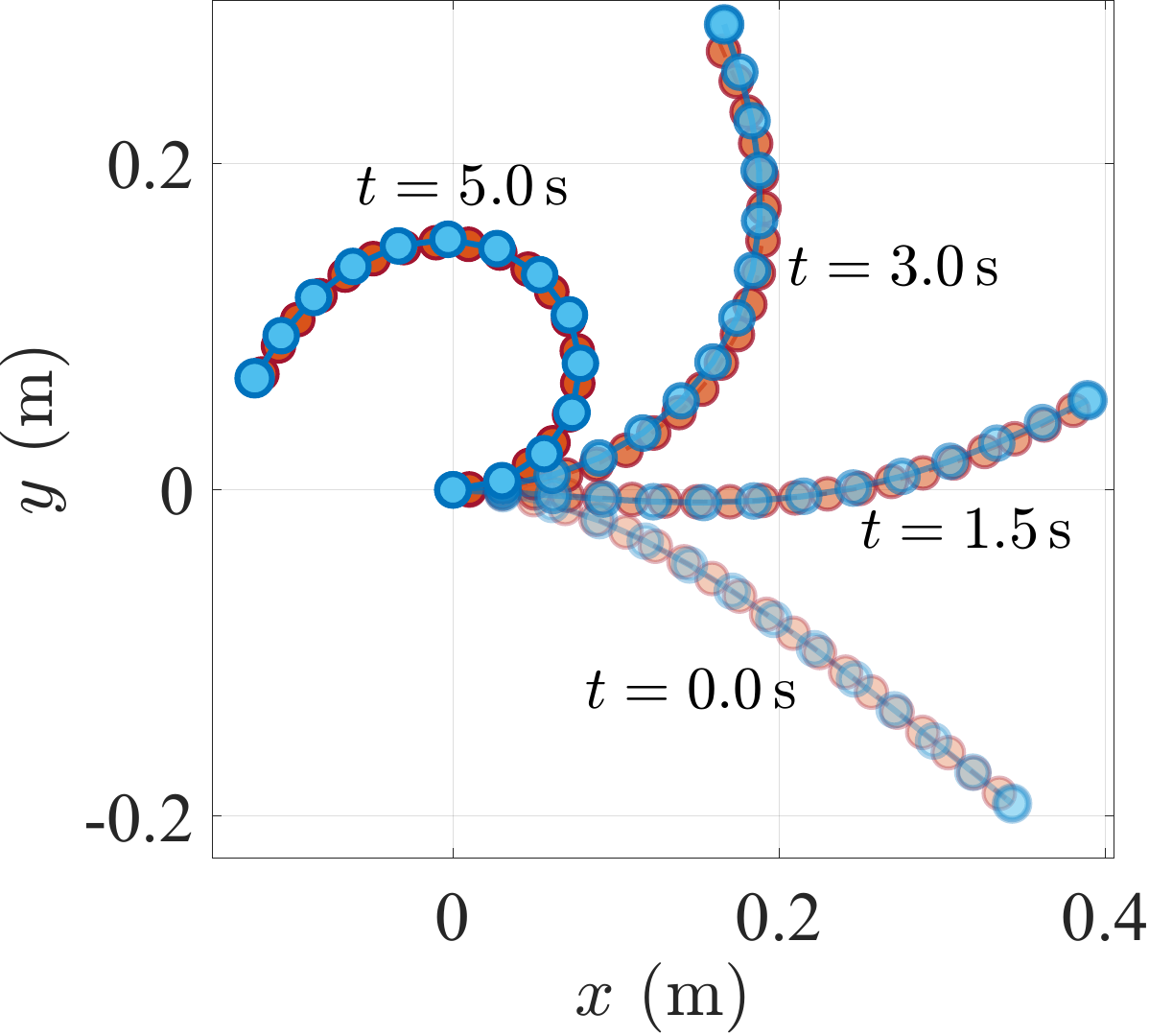}
        \caption{\footnotesize{$\Delta_F(t)=3.16t$}}\label{F_Ws_linear}
    \end{subfigure}
    \begin{subfigure}{0.245\textwidth}
        \includegraphics[width=\linewidth]{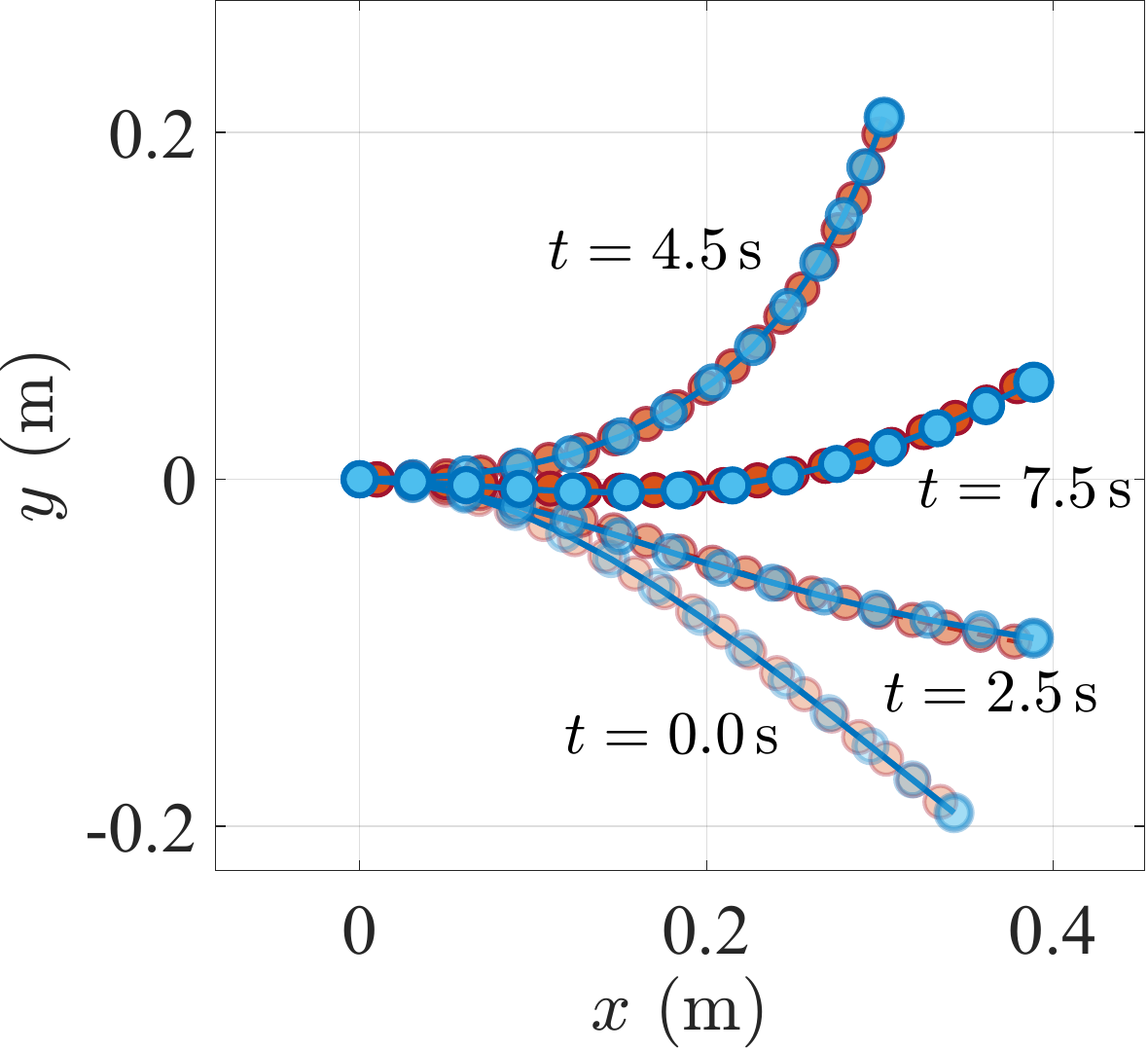}
        \caption{\footnotesize{$\Delta_F(t)\!\!=\!5\!-\!3\sin[2\pi(t-1)]$}}\label{F_Ws_sin}
    \end{subfigure}
    \begin{subfigure}{0.245\textwidth}
        \includegraphics[width=\linewidth]{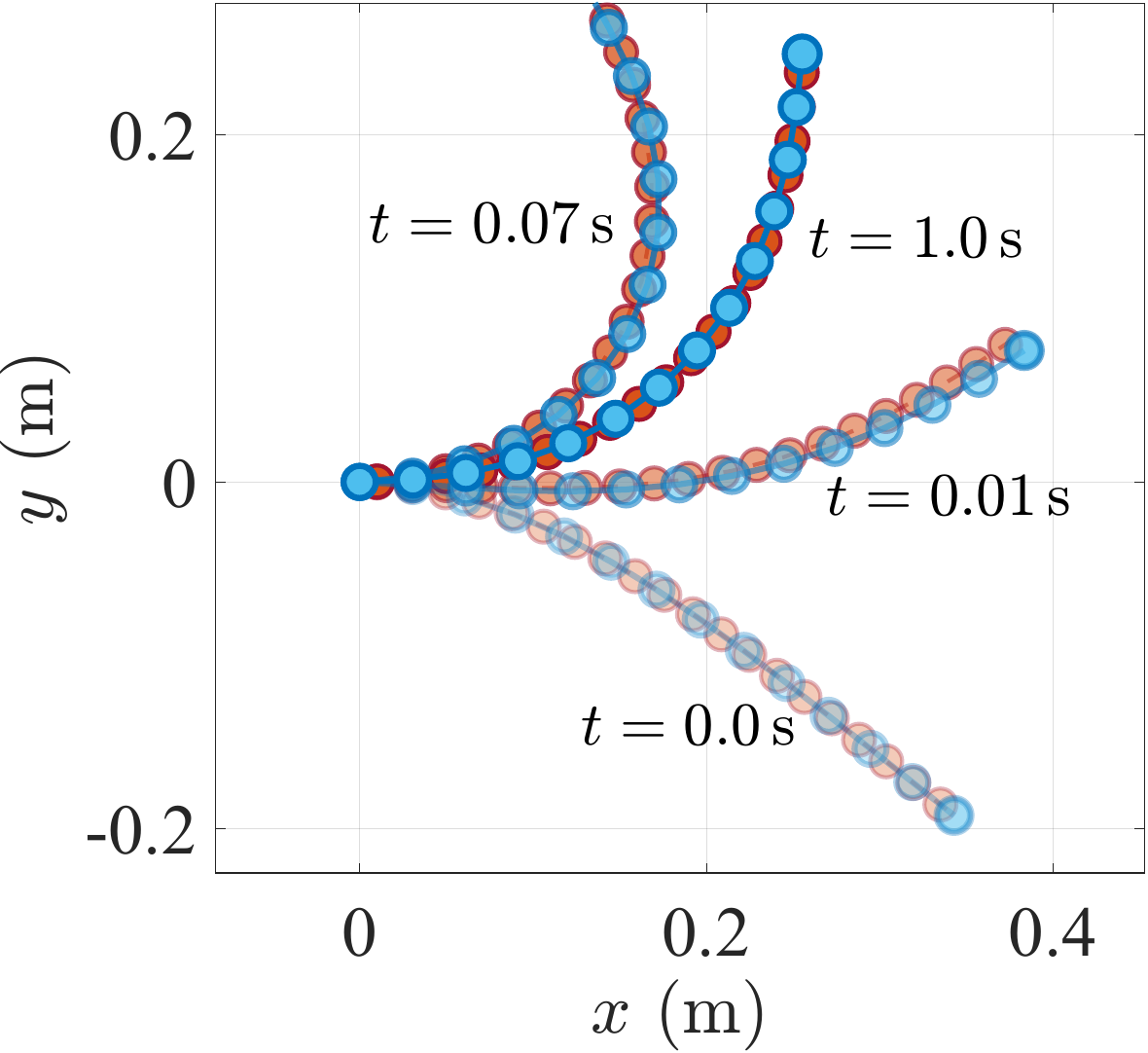}
        \caption{\footnotesize{$\Delta_F(t)=8u(t)$}}\label{F_Ws_step}
    \end{subfigure}
    \begin{subfigure}{0.245\textwidth}
 \includegraphics[width=\linewidth]{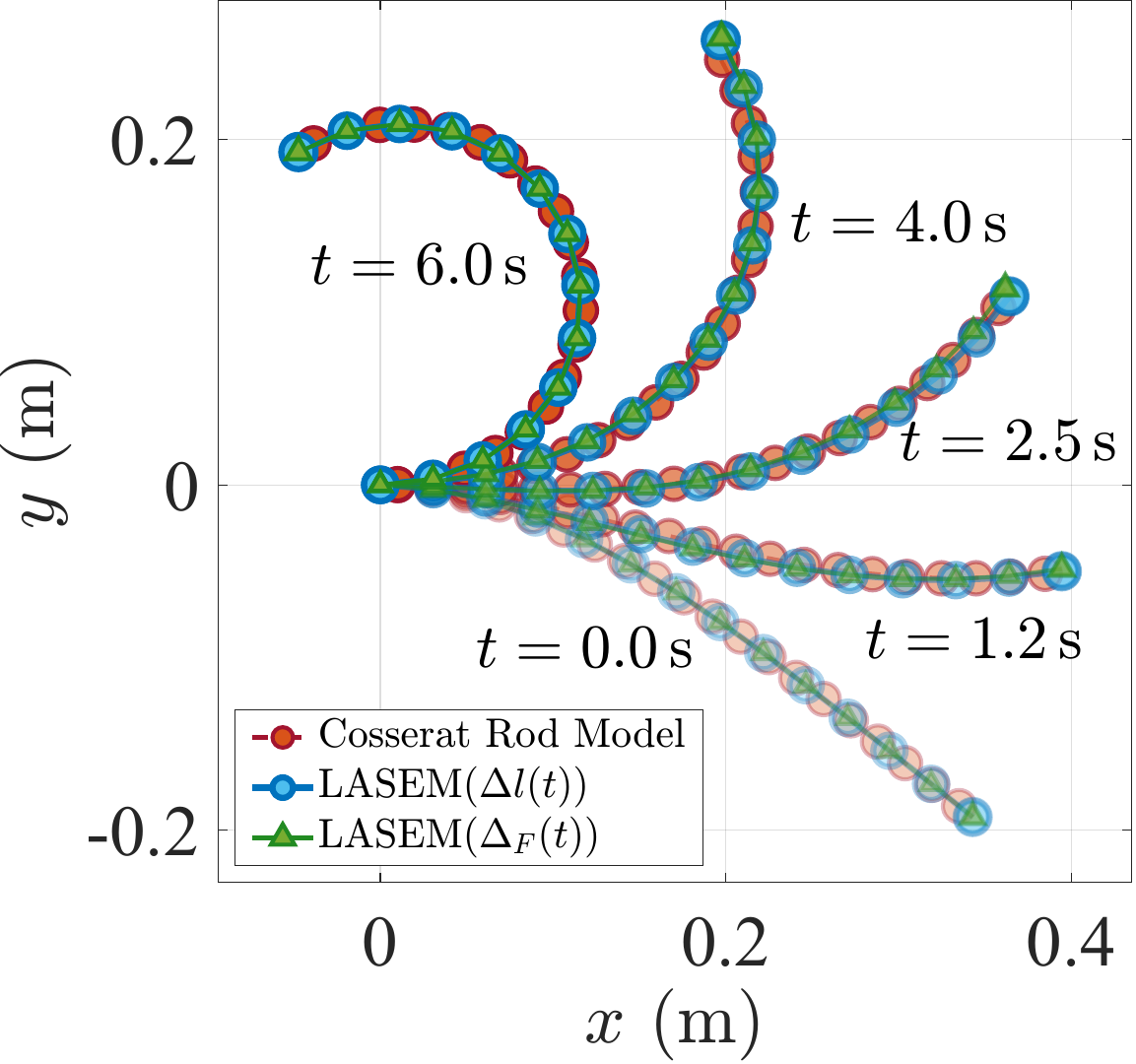}
        \caption{{\footnotesize{$\Delta l(t)=\left|0.048(t-1)\right|$}}}\label{fig:L_con}
    \end{subfigure}
    \begin{subfigure}{0.245\textwidth}
\includegraphics[width=\linewidth]{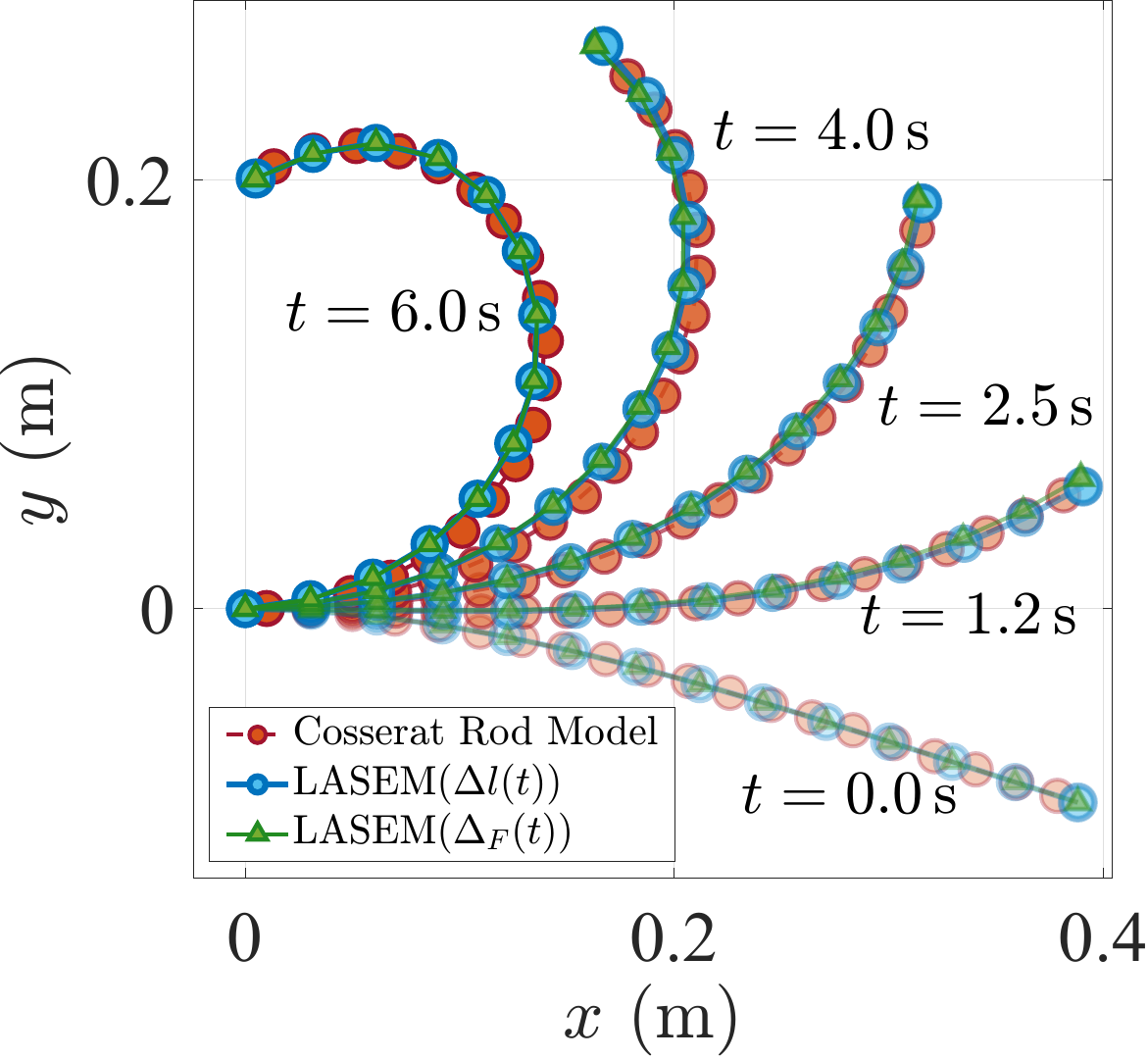}
        \caption{{\footnotesize{$\Delta l(t)=\left|0.048(t-1)\right|$}}}\label{fig:L_Ds}
    \end{subfigure}
    \begin{subfigure}{0.2465\textwidth}
        \includegraphics[width=\linewidth]{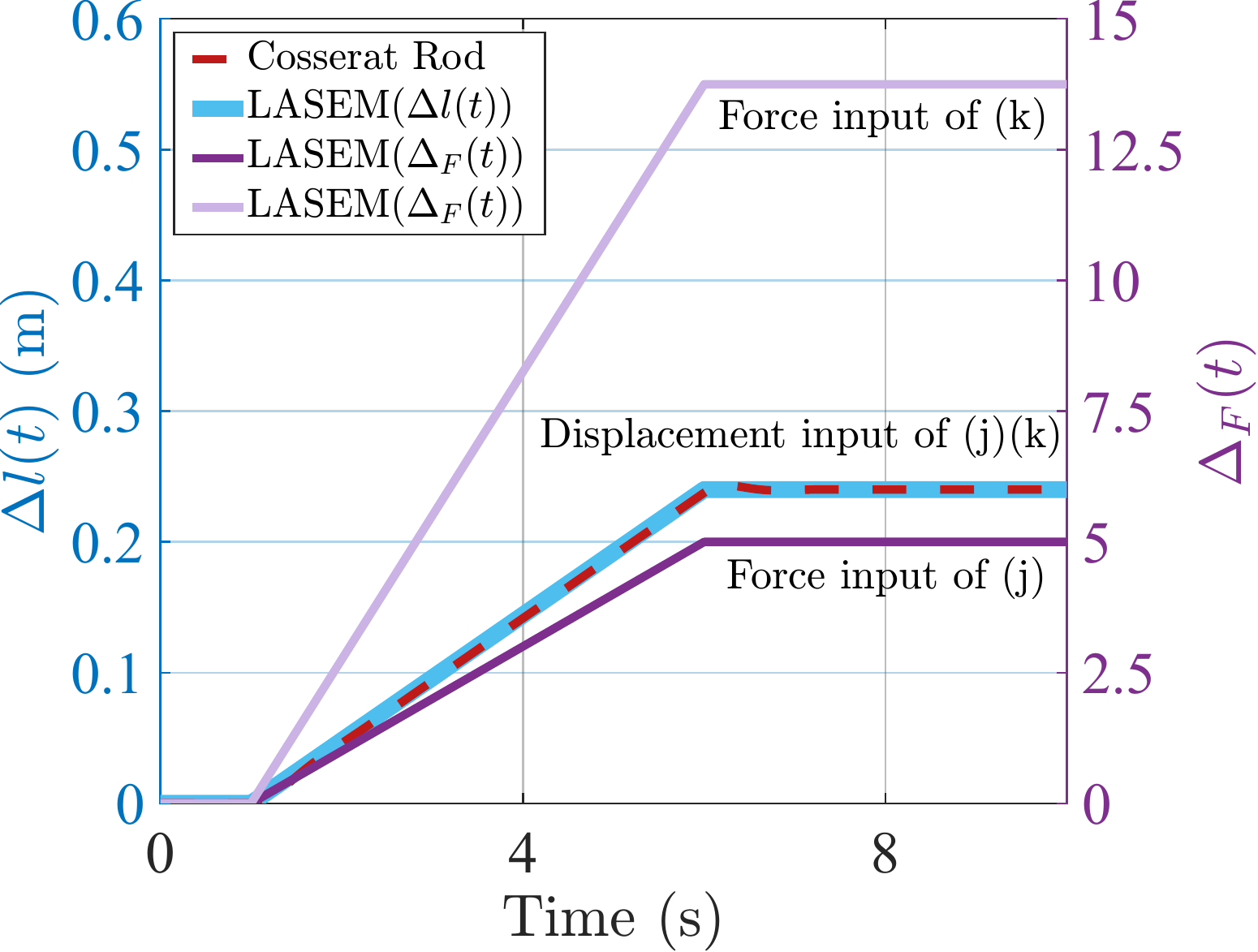}
        \caption{{\footnotesize{Displacement and force of (j) (k)}}}\label{L_input}
    \end{subfigure}
    \caption{\small{Comparison of dynamic responses between LASEM and the Cosserat rod theory. Subfigures (\ref{F_con_linear})–(\ref{F_con_step}) correspond to classic designs (\textbf{Case 1}), (\ref{F_Ds_linear})–(\ref{F_Ds_step}) to nonuniform geometries (\textbf{Case 2}), (\ref{F_Ws_linear})–(\ref{F_Ws_step}) to arbitrary cable routings (\textbf{Case 3}), and (\ref{fig:L_con})–(\ref{fig:L_Ds}) to displacement-input (\textbf{Case 4}), (\ref{L_input}) Displacement {and force} input for cases (\ref{fig:L_con}) and (\ref{fig:L_Ds}). $u(t)$ denotes a step function. As with other dynamic models, the GVS formulation operates under force-input actuation \cite{mathew2025analytical}. 
A controller is therefore required to track the prescribed cable displacement and generate the corresponding 
force inputs for GVS models. {\textbf{Note:} Force reconstruction: simulation-only validation.}}}
    \label{fig:sim_all}
\end{figure*}

\subsection{Case Studies and Settings}

All simulations are conducted using the baseline physical and geometric parameters listed in 
Table~\ref{tab:exp_sim_params}. To evaluate the generality of the proposed LASEM framework against Cosserat rod formulation, four 
representative cases are examined, each modifying one aspect of the structure, routing, or actuation 
while keeping all other parameters unchanged. {All simulations use a fixed time step of $3\times10^{-4}\,\mathrm{s}$. }The simulations were executed on a workstation equipped with an Intel Core i9-14900HX.

\begin{itemize}

    \item \textbf{Case 1: Classic design.}
    All parameters follow Table~\ref{tab:exp_sim_params}.  
    The following actuation profiles are tested:
    \begin{itemize}
        \item Linear input: $\Delta_F(t)=t$.
        \item Sinusoidal input: $\Delta_F(t)\!\!=\!1.5\!-\!0.3\sin[2\pi(t-1)]$.
        \item High-speed step input: $\Delta_F(t)=3u(t)$.
    \end{itemize}
    Where $u(t)$ denotes a step function.  
    \item \textbf{Case 2: Nonuniform geometries. }
    All baseline parameters are retained except the geometry, which follows
    \begin{subequations}
    \small
    \begin{align}
        I(s) &= \frac{\pi}{64}\Big[D_0+(D_1-D_0)\frac{s}{L}\Big]^4,\\
        A(s) &= \frac{\pi}{4}\Big[D_0+(D_1-D_0)\frac{s}{L}\Big]^2,
    \end{align}
    \end{subequations}
    where $D_0=0.006,D_1=0.005$ are the base and tip diameters.
    
    Tested actuation profiles:
    \begin{itemize}
        \item \small Linear input: $\Delta_F(t)=2.75t$.
        \item Sinusoidal input: $\Delta_F(t)\!\!=\!5\!-\!3\sin[2\pi(t-1)]$.
        \item High-speed ramp input: $\Delta_F(t)=13.75u(t)$.
    \end{itemize}
    \item \textbf{Case 3: Arbitrary cable routing.}
    The baseline model is used except that the cable spacing varies as
    \begin{equation}
    \small
    \begin{aligned}
        W(s)=0.04 - 0.03\left(\frac{s}{L}\right)^3 .
    \end{aligned}
    \end{equation}
    Tested actuation profiles:
    \begin{itemize}
        \item \small Linear input: $\Delta_F(t)=3.16t$.
        \item Sinusoidal input: $\Delta_F(t)\!\!=\!5\!-\!3\sin[2\pi(t-1)]$.
        \item High-speed step input: $\Delta_F(t)=8u(t)$.
    \end{itemize}

    \item \textbf{Case 4: Displacement-input actuation.}
    Geometry follows Case 1; the actuation variable is tendon displacement $\Delta l(t)$.  
    Tested input profiles:
    \begin{itemize}
        \item \small Linear inputs in classical designs and nonuniform geometries: $\Delta l(t)=0.048(t-1)$.
    \end{itemize}
\end{itemize}
\subsection{Numerical Results and Analysis}
\subsubsection{Model Accuracy}
\begin{table}[t]
	\vspace*{0.25cm} 
	\centering
	\caption{Computational Cost under Different Conditions}\footnotesize
	\label{tab:compute_time}
	\begin{subtable}[t]{\linewidth}
	\centering
	\caption{\textbf{Case 1: Classic Design}}
    \begin{tabular}{l l c c c c}
    	\toprule
    	\multicolumn{1}{l}{Model} 
    	& \multicolumn{1}{c}{Method}
    	& \multicolumn{1}{c}{\begin{tabular}{@{}c@{}} Linear \\ Input  \end{tabular}}
    	& \multicolumn{1}{c}{\begin{tabular}{@{}c@{}} Sinusoidal  \\  Input \end{tabular}}
    	& \multicolumn{1}{c}{\begin{tabular}{@{}c@{}} High-speed   \\ Step Input \end{tabular}}
        & \multicolumn{1}{c}{\begin{tabular}{@{}c@{}} {Strain}   \\ {Order} \end{tabular}}
    	\\ 
    	\midrule
    	
    	\multirow{2}{*}{LASEM}
    	& \multicolumn{1}{c}{SD$^{1^*}$} & 2.662 s & 2.278 s & 1.557 s & NA$^{2^*}$\\
    	& \multicolumn{1}{c}{Galerkin} & 0.213  s & 0.107 s & 0.088 s & 6 \\
    	\midrule
    	
    	\multirow{2}{*}{\begin{tabular}{@{}c@{}} Cosserat   \\ rod \end{tabular}}
    	& \multicolumn{1}{c}{GVS}       &  NC$^{3^*}$ & NC & NC & 6 \\
    	& \multicolumn{1}{c}{D-GVS}     &  0.523 s & 0.425 s & 0.242 s & 6 \\
    	\midrule
        
        \multicolumn{1}{c}{{PCC$^{4^*}$}}
    	& \multicolumn{1}{c}{{--}} & {0.098 s} & {0.132 s} & {0.094 s} & {NA}\\
    	\bottomrule
    \end{tabular}
	\end{subtable}
    
    \vspace{6pt}
	\begin{subtable}[t]{\linewidth}
		\centering
		\caption{\textbf{Case 2: Nonuniform geometries }}
            \begin{tabular}{l l c c c c}
    	\toprule
    	\multicolumn{1}{l}{Model} 
    	& \multicolumn{1}{c}{Method}
    	& \multicolumn{1}{c}{\begin{tabular}{@{}c@{}} Linear \\ Input  \end{tabular}}
    	& \multicolumn{1}{c}{\begin{tabular}{@{}c@{}} Sinusoidal  \\  Input \end{tabular}}
    	& \multicolumn{1}{c}{\begin{tabular}{@{}c@{}} High-speed   \\ Step Input \end{tabular}}
        & \multicolumn{1}{c}{\begin{tabular}{@{}c@{}} Strain   \\ Order \end{tabular}}
    	\\ 
    	\midrule
    	
    	\multirow{2}{*}{LASEM}
    	& \multicolumn{1}{c}{SD} & 2.218 s & 2.022 s & 1.445 s & NA\\
    	& \multicolumn{1}{c}{Galerkin} & 0.142  s & 0.078 s & 0.070 s & 6 \\
    	\midrule
    	
    	\multirow{2}{*}{\begin{tabular}{@{}c@{}} Cosserat   \\ rod \end{tabular}}
    	& \multicolumn{1}{c}{GVS}       &  NC & NC & NC & 6 \\
    	& \multicolumn{1}{c}{D-GVS}     &  0.209 s & 0.276 s & 0.130 s & 6 \\
    	
    	\bottomrule
    \end{tabular}
	\end{subtable}
    
    \vspace{6pt}
	\begin{subtable}[t]{\linewidth}
		\centering
		\caption{\textbf{Case 3: Arbitrary Cable Routing}}
		    \begin{tabular}{l l c c c c}
    	\toprule
    	\multicolumn{1}{l}{Model} 
    	& \multicolumn{1}{c}{Method}
    	& \multicolumn{1}{c}{\begin{tabular}{@{}c@{}} Linear \\ Input  \end{tabular}}
    	& \multicolumn{1}{c}{\begin{tabular}{@{}c@{}} Sinusoidal  \\  Input \end{tabular}}
    	& \multicolumn{1}{c}{\begin{tabular}{@{}c@{}} High-speed   \\ Step Input \end{tabular}}
        & \multicolumn{1}{c}{\begin{tabular}{@{}c@{}} Strain   \\ Order \end{tabular}}
    	\\ 
    	\midrule
    	
    	\multirow{2}{*}{LASEM}
    	& \multicolumn{1}{c}{SD} & 2.743 s & 2.246 s & 1.270 s & NA\\
    	& \multicolumn{1}{c}{Galerkin} & 0.113 s & 0.121 s & 0.081 s & 6 \\
    	\midrule
    	
    	\multirow{2}{*}{\begin{tabular}{@{}c@{}} Cosserat   \\ rod \end{tabular}}
    	& \multicolumn{1}{c}{GVS}       &  NC & NC & NC & 6 \\
    	& \multicolumn{1}{c}{D-GVS}     &  0.406 s & 0.331 s & 0.144 s & 6 \\
    	
    	\bottomrule
    \end{tabular}
	\end{subtable}   
\caption*{ \scriptsize{\textbf{Note:} 1$^{*}$ Spatial discretization (SD) is a standard technique for solving partial differential equations \cite{leveque2007finite}.  2$^{*}$ NA denotes not applicable. Unlike the strain-order formulation, SD discretizes the system in space. 3$^{*}$ {NC denotes computation time exceeding 2 hours. }{4$^{*}$ For PCC, strain order is not applicable; the baseline uses six constant-curvature sections. Code reference: \url{https://github.com/tarquasso/3d_soft_trunk_contact}} }}
\end{table}
Numerical simulations were conducted using both the force-input model in Eq.~\eqref{dynamic_in_F} 
and the displacement-input model in Eq.~\eqref{dynamic_in}. Representative results for all test cases 
are shown in Fig.~\ref{fig:sim_all}. Across all input profiles in Cases~1–4, the deformation shapes 
predicted by LASEM (SD or Galerkin) generally agree with those obtained from the Cosserat-rod models 
(GVS or D-GVS), with only modest deviations. {The PCC-based dynamic approximation \cite{katzschmann2019dynamic} included in Fig.~\ref{fig:sim_all} provides an additional reduced-order reference.} These differences arise from both modeling structure and 
numerical treatment: LASEM is governed by a single Euler–moment-balance PDE that implicitly enforces 
force balance, whereas Cosserat-rod formulations solve coupled force and moment equations, naturally 
leading to different discretization characteristics. 

\subsubsection{Computational Expense}
{Table~\ref{tab:compute_time} reports the computation times for all methods. LASEM-Galerkin is faster than PCC and SD \cite{leveque2007finite}. Although PCC is a reduced-order model \cite{katzschmann2019dynamic}, its Lagrangian dynamics still require repeated numerical evaluation of configuration-dependent terms. GVS likewise incurs numerical Jacobian approximation without analytical derivatives, whereas D-GVS and LASEM-Galerkin use analytical derivatives for faster computation. LASEM-Galerkin further achieves lower runtime than D-GVS. The relative efficiency is quantified by the speedup metric}
\begin{equation}
\small
\begin{aligned}
    \textbf{Speedup}
    = \frac{T_{\mathrm{D\!-\!GVS}} - T_{\mathrm{LASEM\!-\!Galerkin}}}
           {T_{\mathrm{D\!-\!GVS}}},
\end{aligned}
\end{equation}
\begin{figure}[b] 
	\vspace*{0.22cm} 
    \centering
\includegraphics[width=0.75\linewidth]{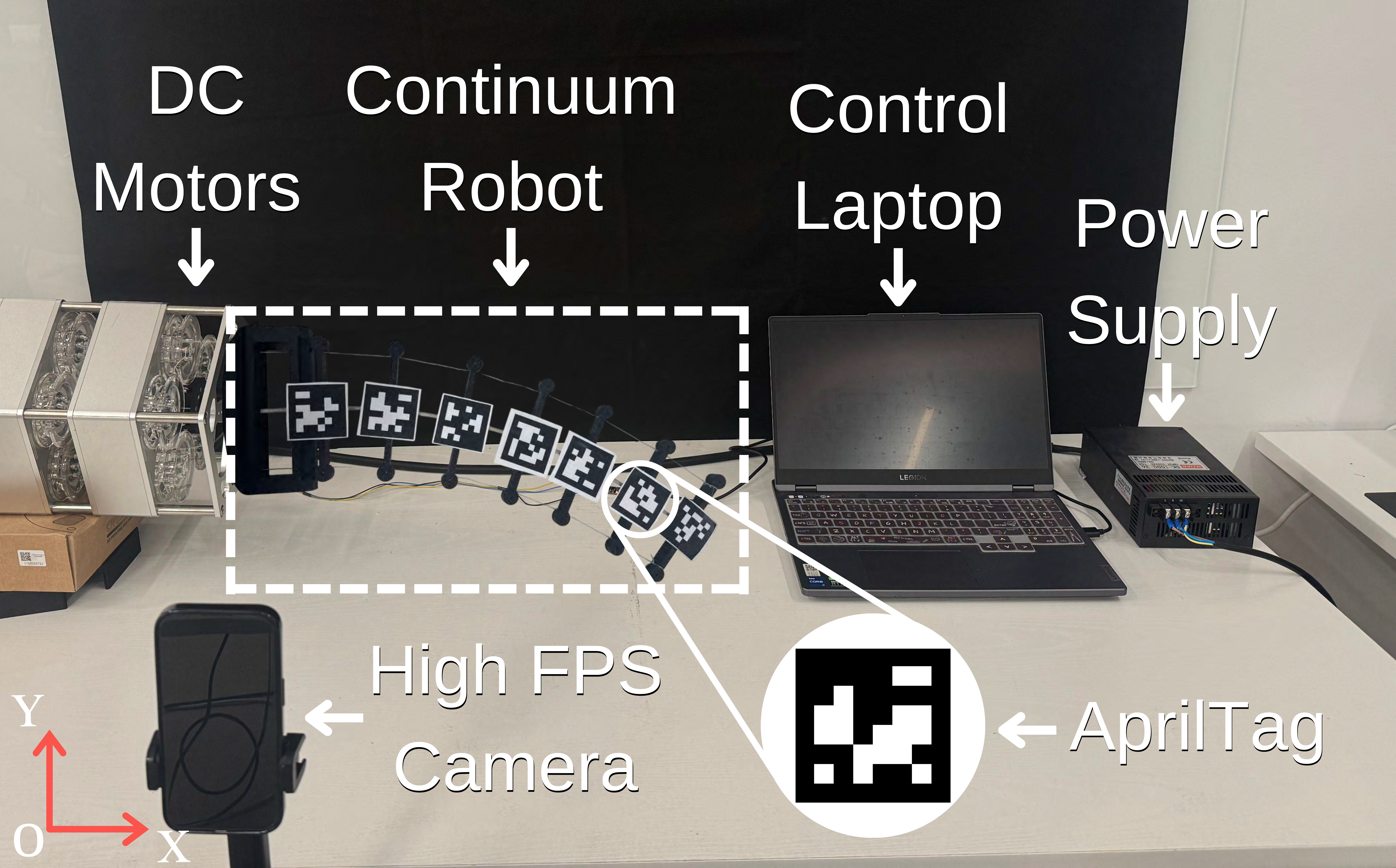}
    \caption{Experimental Setup
}
    \label{fig:exp}
\end{figure}
which measures the normalized reduction in runtime. Using this metric, the Galerkin-based LASEM
implementation achieves an average \textbf{62.3\% Speedup} across all test cases, outperforming
D-GVS while maintaining high modeling accuracy. Therefore, the improved computational efficiency makes the proposed model more suitable for real-time applications, such as model predictive control.
\begin{figure*}[t]
	\vspace*{0.2cm} 
    \centering
    \begin{subfigure}{0.235\textwidth}
        \includegraphics[width=\linewidth]{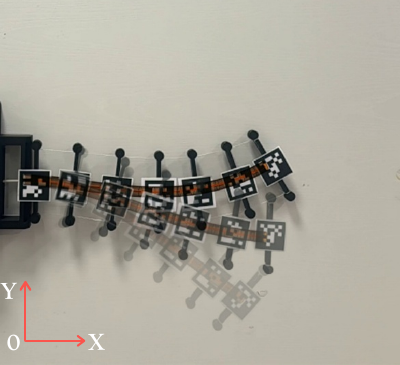}
        \caption{\footnotesize{Case A: Exp vs. Sim}}\label{exp_linear_PNG}
    \end{subfigure}
    \begin{subfigure}{0.235\textwidth}
        \includegraphics[width=\linewidth]{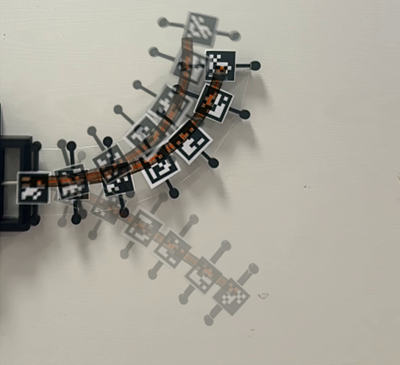}
        \caption{\footnotesize{Case B}}\label{exp_step}
    \end{subfigure}
    \begin{subfigure}{0.235\textwidth}
        \includegraphics[width=\linewidth]{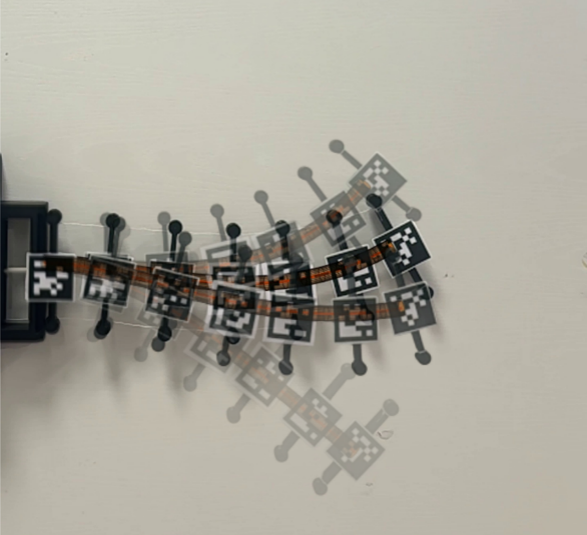}\label{exp_sin}
        \caption{\footnotesize{Case C}}
    \end{subfigure}
    \begin{subfigure}{0.235\textwidth}
        \includegraphics[width=\linewidth]{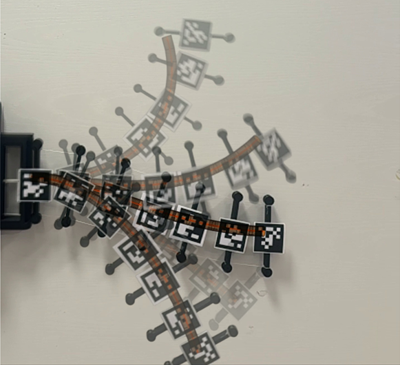}
        \caption{\footnotesize{Case D}}\label{exp_zd}
    \end{subfigure}
    \begin{subfigure}{0.24\textwidth}
        \includegraphics[width=0.96\linewidth]{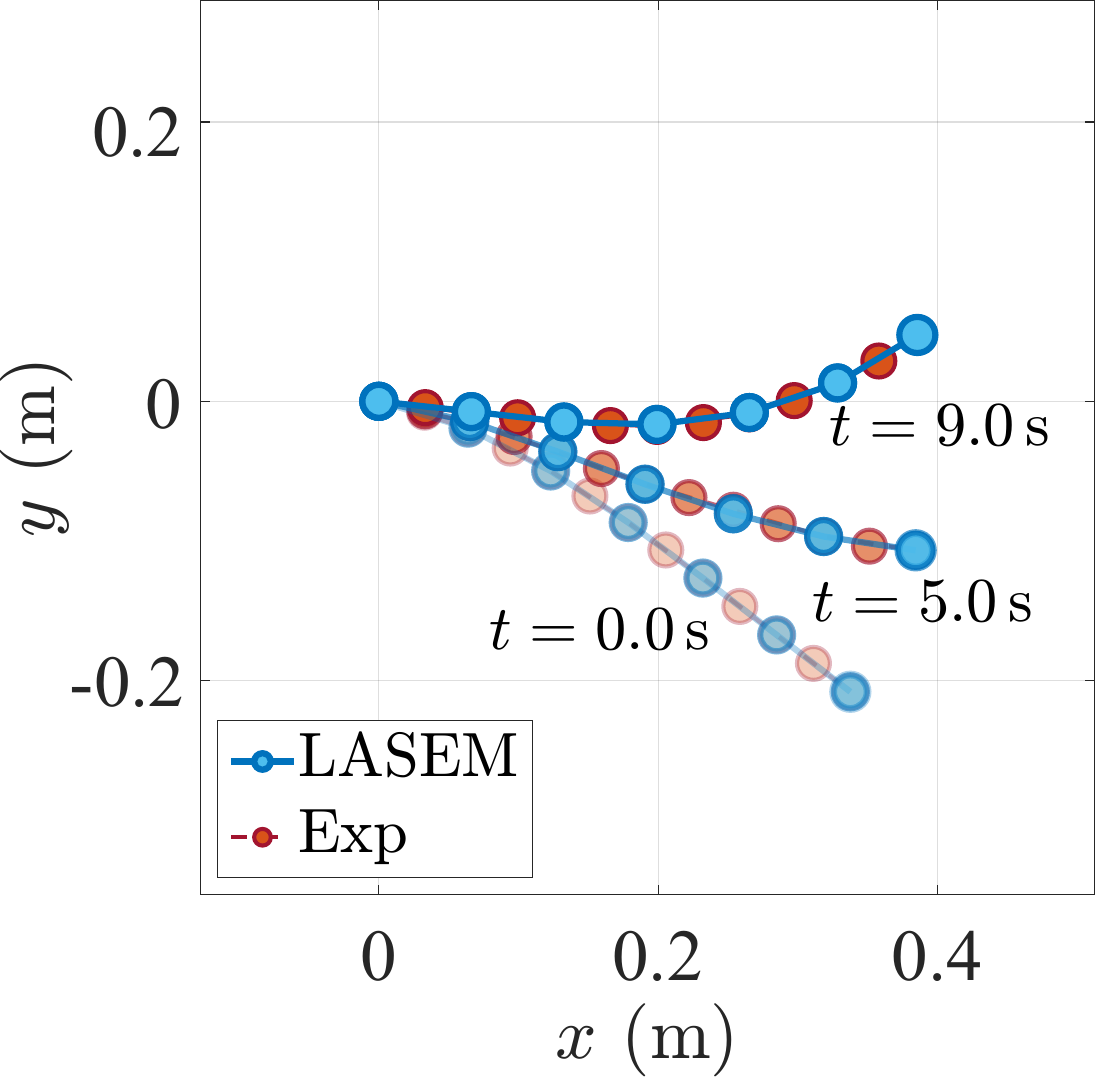}
        \caption{\footnotesize{Case A: Shape trajectory }}\label{fig4:linear}
    \end{subfigure}
    \begin{subfigure}{0.24\textwidth}
    \includegraphics[width=\linewidth]{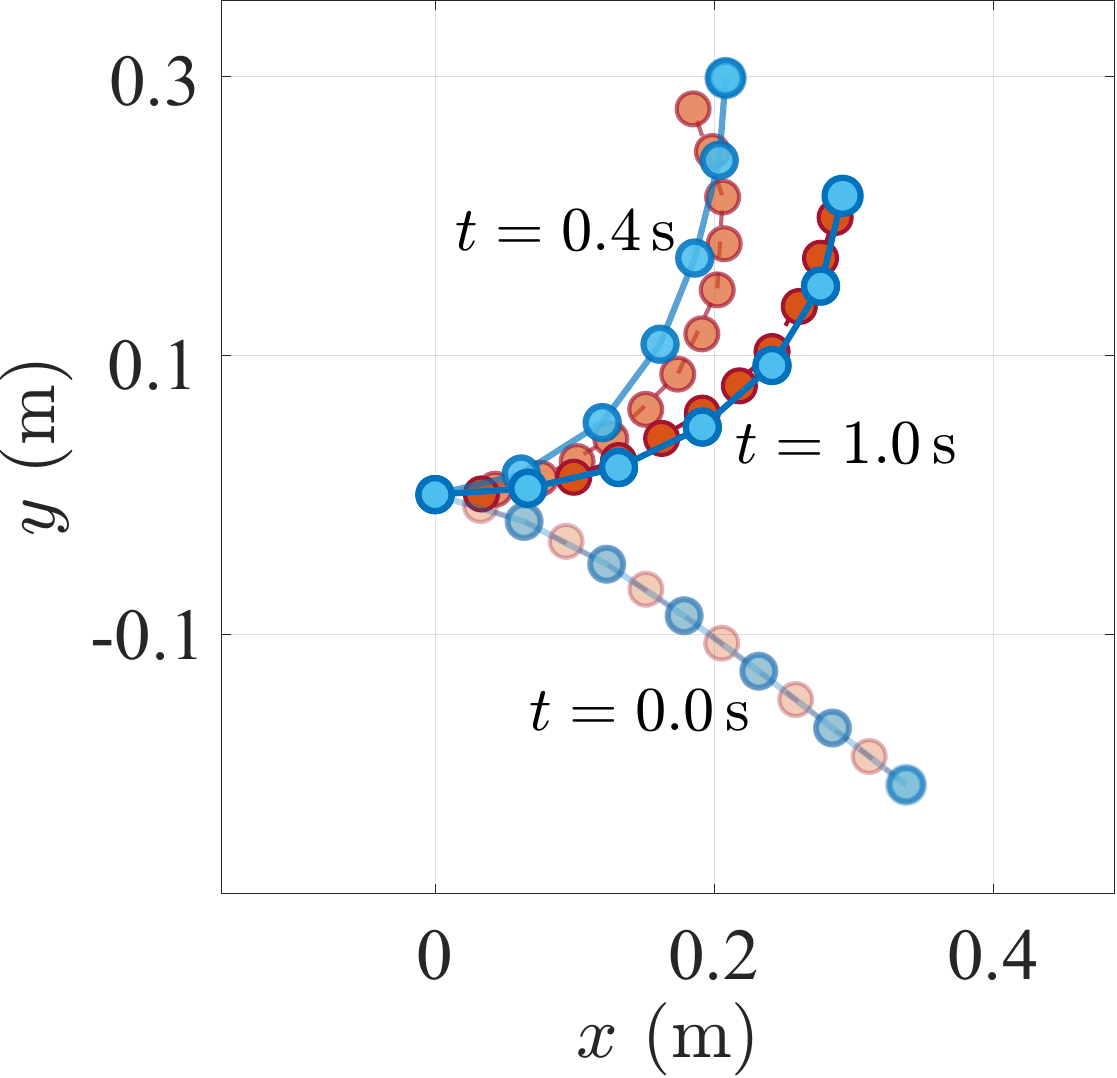}
        \caption{\footnotesize{Case B}}\label{fig4:step}
    \end{subfigure}
    \begin{subfigure}{0.24\textwidth}
    \includegraphics[width=\linewidth]{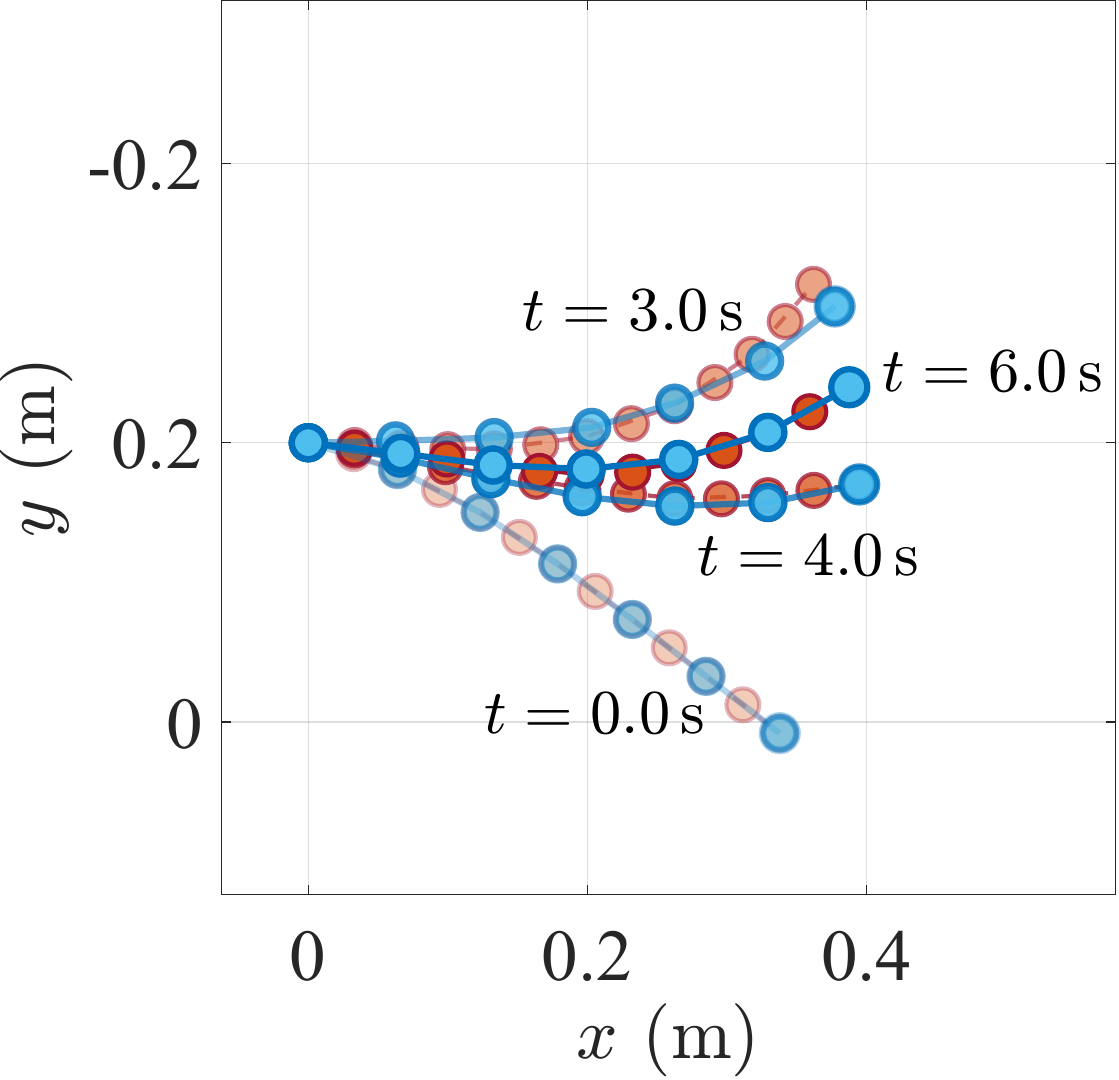}
        \caption{\footnotesize{Case C}}\label{fig4:sin}
    \end{subfigure}
    \begin{subfigure}{0.24\textwidth}
        \includegraphics[width=\linewidth]{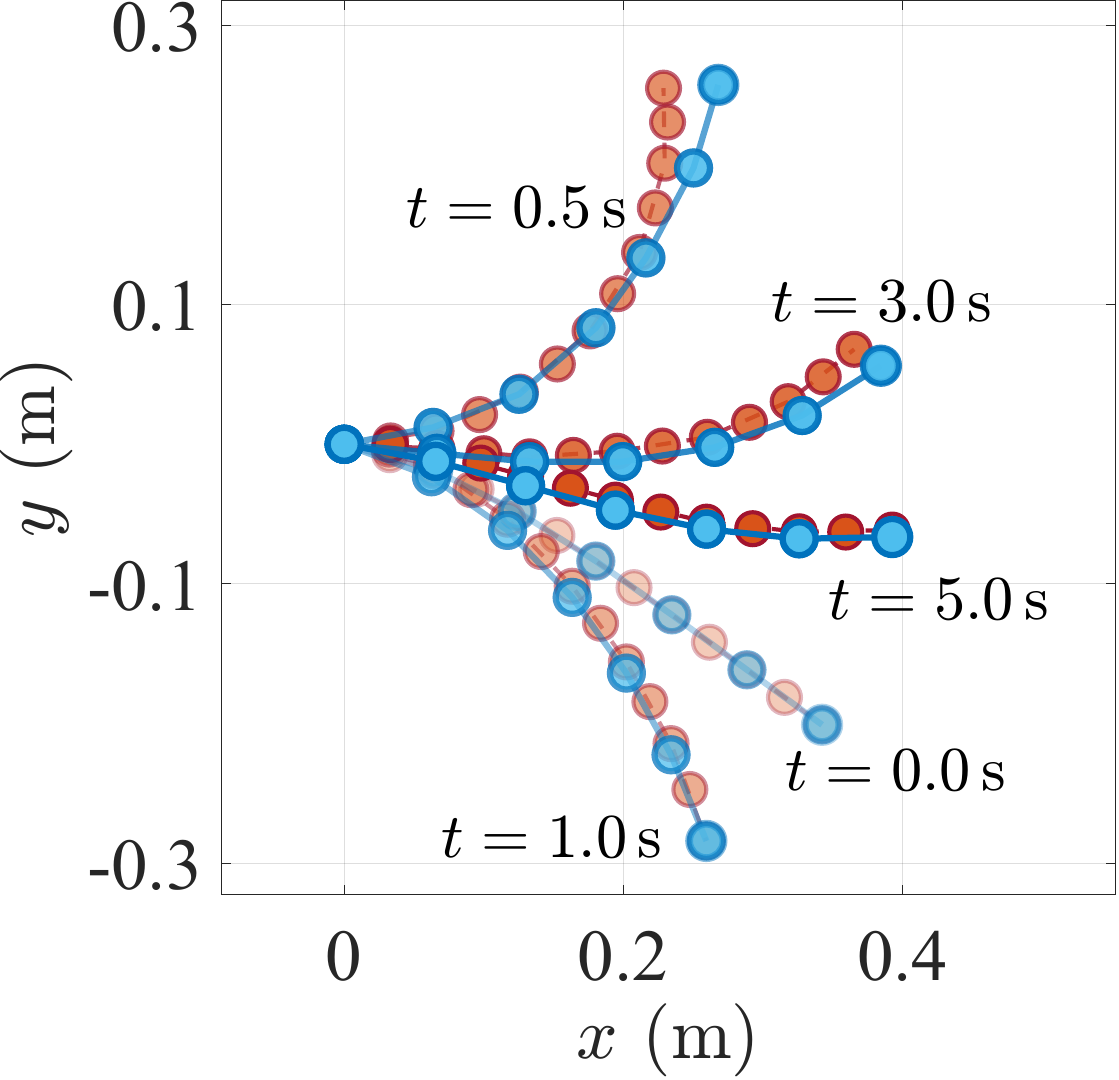}
        \caption{\footnotesize{Case D}}\label{fig4:zd}
    \end{subfigure}
    \begin{subfigure}{0.246\textwidth}
        \includegraphics[width=\linewidth]{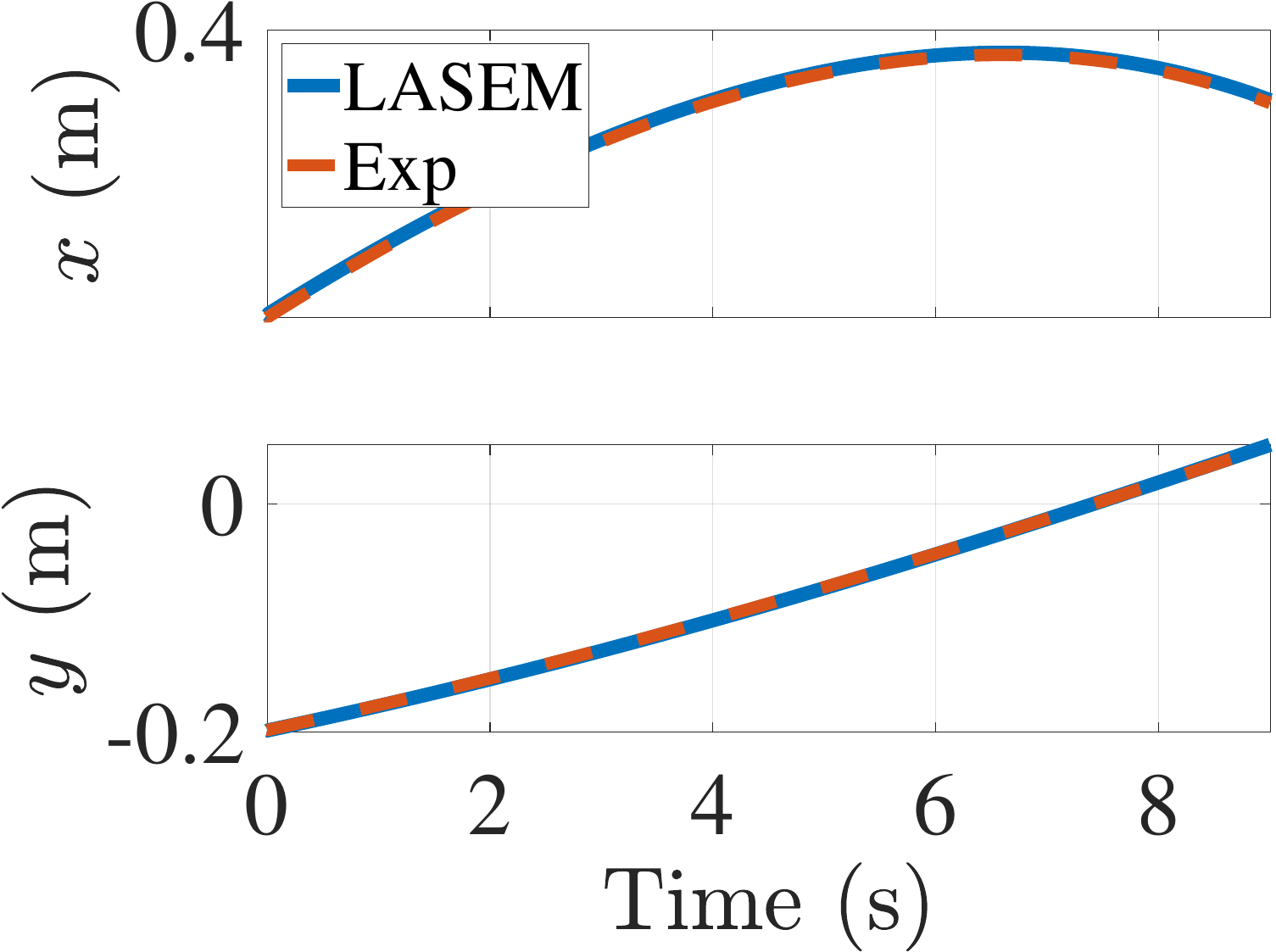}
        \caption{\footnotesize{Case A: End-effector trajectory}}\label{fig4:linear_xy}
    \end{subfigure}
    \begin{subfigure}{0.245\textwidth}
 \includegraphics[width=\linewidth]{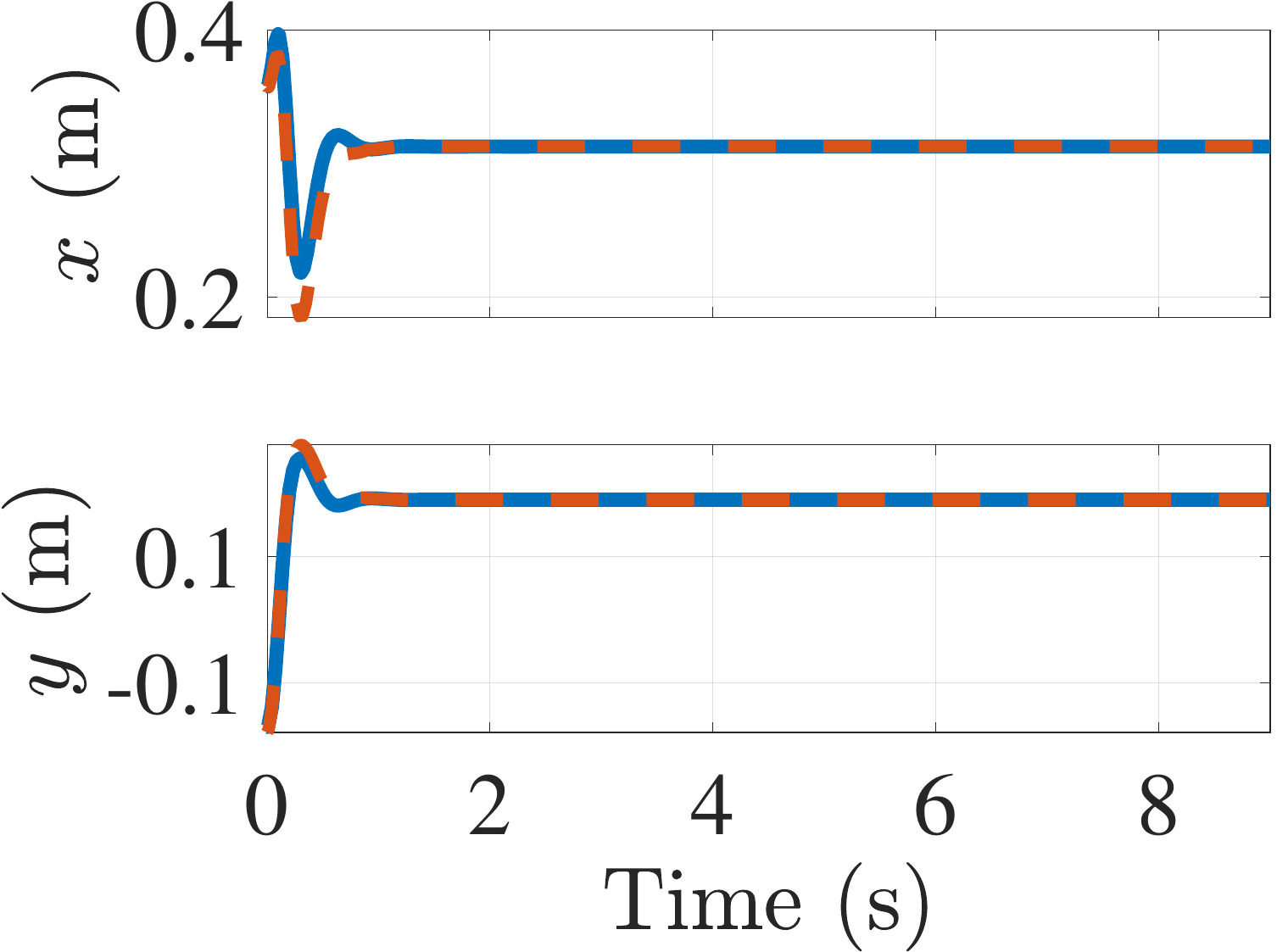}
        \caption{\footnotesize{Case B}}\label{step_xy}
    \end{subfigure}
    \begin{subfigure}{0.245\textwidth}
\includegraphics[width=1.15\linewidth]{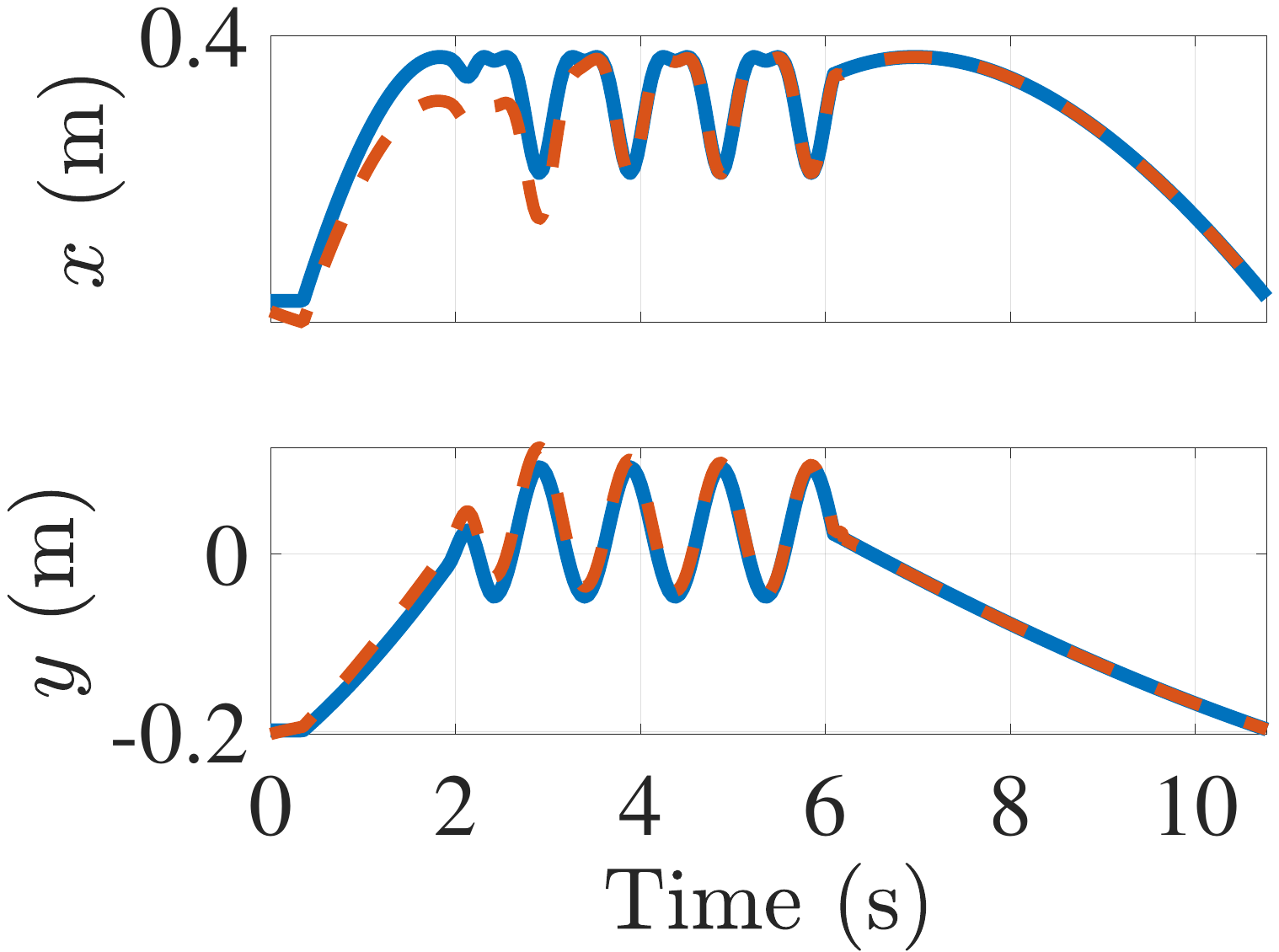}
        \caption{\footnotesize{Case C}}\label{sin_xy}
    \end{subfigure}
    \begin{subfigure}{0.245\textwidth}
        \includegraphics[width=1.15\linewidth]{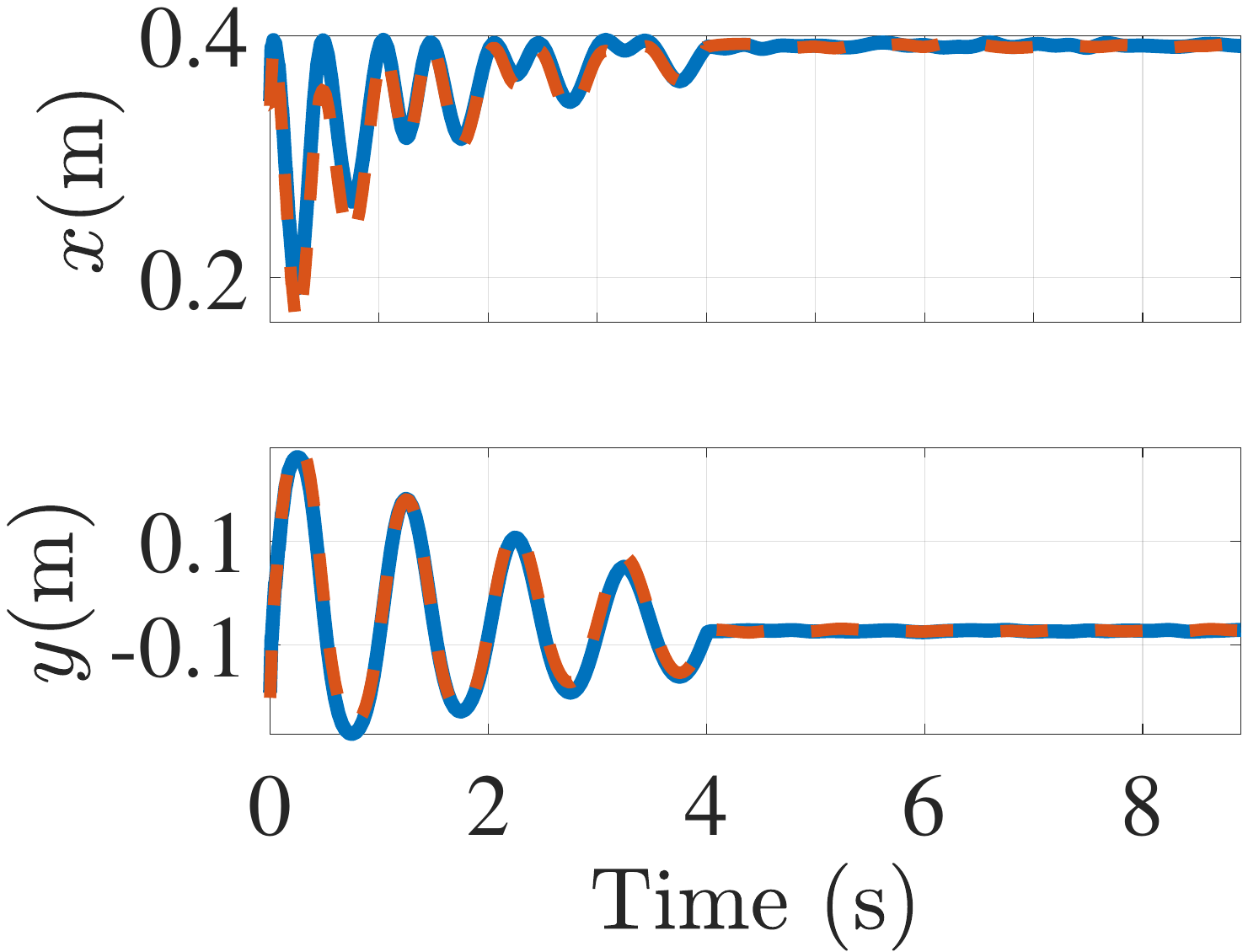}
        \caption{\footnotesize{Case D}}\label{fig4:zd_xy}
    \end{subfigure}
    \caption{\small{Comparison between experimental measurements and simulation results across four test cases, including experiment vs. simulation comparisons (\ref{exp_linear_PNG})-(\ref{exp_zd}), backbone-shape trajectory comparisons (\ref{fig4:linear})-(\ref{fig4:zd}), and end-effector trajectory comparisons (\ref{fig4:linear_xy})-(\ref{fig4:zd_xy}).}}
    \label{fig:exp_result}
\end{figure*}


\section{Experimental Validation}
Physical experiments are conducted under displacement actuation, consistent with LASEM’s
displacement-input formulation in which the cable displacement $\Delta l(t)$ serves as the
actuation input. This native compatibility \eqref{dynamic_in} enables direct
comparison between simulation and experiment.

\subsection{Experimental Setup}

As shown in Fig.~\ref{fig:exp}, the CDCR is actuated by two DC motors prescribing two cable
displacements, and its posture is captured using a vision-based tracking system with planar AprilTag
markers mounted along the backbone. Because the experiments involve rapid motions, a high-speed
camera operating at 160~Hz is employed to ensure sufficient temporal resolution for accurate motion
reconstruction. The geometric, material, and design parameters follow Table~\ref{tab:exp_sim_params};
note that the parallel-axis cable routing is physically implemented through seven rigid disks that
discretize the tendon paths along the backbone (Fig.~\ref{fig:exp}).

\subsection{Case Studies}

To evaluate the dynamic performance of the proposed LASEM framework for dynamic modeling under different displacement actuation patterns,
four representative cable-displacement inputs are tested.  
The input profiles and their mathematical expressions are summarized below.

\begin{itemize}
    \item \textbf{Case A: \small Slow linear ramp: $\Delta l(t) = 0.008 t, 0\le t\le 9.$}
    \item \textbf{Case B: \small Step input: $\Delta l(t) = 0.14u(t), 0\le t\le 9.$}
    \item \textbf{Case C: \small Quick linear ramp and sinusoidal excitation:}
    \begin{equation*}
    \small
    \begin{aligned}
        \Delta l(t) =
        \begin{cases}
            0.04  t, & 0 \le t < 2,\\[4pt]
            0.08  + 0.018 \sin\!\big[2\pi (t - 2)\big], & 2\le t < 6,\\[4pt]
            0.24 - \frac{0.08}{3}t, & 6 \le t \le 11. 
        \end{cases}
    \end{aligned}
    \end{equation*}
    \item \textbf{Case D: \small Exponentially decaying sinusoidal input:}
    \begin{equation*}
    \small
    \begin{aligned}
    \Delta l(t) = 
    \begin{cases}
        0.046-0.12\, e^{-0.35 t}\, \sin(2\pi t), & 0 \le t < 4,\\[4pt]
        0.046 & 4\le t \le 9.
        \end{cases}
    \end{aligned}
    \end{equation*}
\end{itemize}
\subsection{Experimental Results and Analysis}

The comparisons between the measured configurations obtained from the high-speed camera and the
simulated results are shown in Figs.~(\ref{exp_linear_PNG})–(\ref{exp_zd}), demonstrating close
agreement in both deformation profiles and transient responses. Prediction accuracy is further
assessed through backbone-shape trajectories in Figs.~(\ref{fig4:linear})–(\ref{fig4:zd}) and
end-effector trajectories in Figs.~(\ref{fig4:linear_xy})–(\ref{fig4:zd_xy}). Across all input modes
from \textbf{Case~A} to \textbf{Case~D}, the simulated responses produced by LASEM closely track the
time-varying motions of the physical robot. The remaining discrepancies can be attributed to factors
such as parameter-identification uncertainty and structural imperfections in the hardware. 


\section{Conclusions}
This work introduced the Lightweight Actuation-Space Energy Modeling (LASEM) framework for dynamic
modeling of CDCRs. By formulating actuation potential energy directly in
actuation space, LASEM yields a single Euler–moment–balance PDE that implicitly accounts for force
balance and avoids explicit cable–backbone contact-force computation. The framework accommodates both
force- and displacement-input actuation and applies naturally to different CDCR structures.
With the derived analytical derivatives, LASEM achieves an average \textbf{62.3\% }computational speedup over existing
real-time dynamic models. Numerical and experimental validations show that LASEM reproduces the
manipulator’s deformation and dynamic responses with high fidelity and computational efficiency.
{Future work will focus on three-dimensional dynamic modeling, probabilistic modeling, and experimental validation of the proposed force-reconstruction framework.}

\bibliographystyle{IEEEtran}
\bibliography{reference}


\newpage

\vfill

\end{document}